\documentclass[letterpaper, 10 pt, conference]{ieeeconf}  

\IEEEoverridecommandlockouts                              

\usepackage{graphics} 
\usepackage{epsfig} 
\usepackage{mathptmx} 
\usepackage{amsmath} 
\usepackage{amssymb}  
\usepackage{graphicx}
\usepackage{bm}
\usepackage{eucal}
\usepackage{todonotes}[inline]  

\usepackage{float}
\usepackage[ruled,vlined]{algorithm2e}
\usepackage{pgfplotstable}
\usepackage{pgfplots}
\usepackage{tikz}
\usetikzlibrary[pgfplots.groupplots]
\usepgfplotslibrary{fillbetween}
\pgfplotsset{compat=1.5}
\usepackage{subcaption}
\usepackage{caption}
\usepackage{xcolor}
\usepackage{enumerate}
\usepackage{makecell}
\usepackage{colortbl}

\usepackage{graphicx}
\usepackage{tabularx,ragged2e,booktabs}
\usepackage{url}
\usepackage{siunitx}
\usepackage{rotating}
\usepackage{amssymb}
\usepackage{tabularx}
\usepackage{pgfplots}
\usepackage{csvsimple}
\usepackage{multirow}
\usepackage[noadjust]{cite}
\usepackage{float}
\usepackage{graphicx}
\usepackage{booktabs}
\usepackage{wrapfig}
\usepackage[font=footnotesize]{caption}

\DeclareMathOperator*{\argmax}{arg\,max}

\title{\LARGE \bf
\textit{Touch if it's transparent!} \\ ACTOR: Active Tactile-based Category-Level \\Transparent Object Reconstruction
}

\author {Prajval Kumar Murali, Bernd Porr, and Mohsen Kaboli
\thanks{P.K.Murali and M.Kaboli are with the BMW Group, Munich Germany. 
e-mail: name.surname@bmwgroup.com}%
\thanks{P.K. Murali and B. Porr are with the University of Glasgow, Scotland}%
\thanks{M. Kaboli is with the Donders Institute for Brain and Cognition, Radboud University, Netherlands }%
\thanks{Funded in part by the BMW Group, EU H2020 INTUITIVE under Grant ID 861166 and  EU Horizon PHASTRAC under Grant ID 101092096.}}
\begin{document}
\bstctlcite{IEEEexample:BSTcontrol}
\maketitle
\thispagestyle{empty}
\pagestyle{empty}

\begin{abstract}
Accurate shape reconstruction of transparent objects is a challenging task due to their non-Lambertian surfaces and yet necessary for robots for accurate pose perception and safe manipulation. As vision-based sensing can produce erroneous measurements for transparent objects, the tactile modality is not sensitive to object transparency and can be used for reconstructing the object's shape. We propose \textit{ACTOR}, a novel framework for \textit{AC}tive tactile-based category-level \textit{T}ransparent \textit{O}bject \textit{R}econstruction. ACTOR leverages large datasets of synthetic object with our proposed self-supervised learning approach for object shape reconstruction as the collection of real-world tactile data is prohibitively expensive. ACTOR can be used during inference with tactile data from category-level unknown transparent objects for reconstruction. Furthermore, we propose an active-tactile object exploration strategy as probing every part of the object surface can be sample inefficient. We also demonstrate tactile-based category-level object pose estimation task using ACTOR. We perform an extensive evaluation of our proposed methodology with real-world robotic experiments with comprehensive comparison studies with state-of-the-art approaches. Our proposed method outperforms these approaches in terms of tactile-based object reconstruction and object pose estimation.
\end{abstract}

\section{INTRODUCTION}
\label{sec:introduction}

Transparent objects such as cups, glasses, and bottles are ubiquitous around us and if robots are expected to work in unstructured scenarios such as household environments, it is essential to recognize and safely manipulate transparent objects. Reconstruction of the object shape is critical for detecting and identifying its pose and safely manipulating it~\cite{Qiang-TRO-2020}. While this is straightforward for opaque objects with off-the-shelf vision sensors, such sensors produce unreliable and erroneous data with transparent objects due to their non-Lambertian surfaces. Sophisticated custom calibrated setups with specialized scanners or modifying the transparent surface of objects are often necessary for accurate reconstruction~\cite{ihrke2010transparent,li2020through}. This is impractical for on-the-fly reconstruction of arbitrary unknown objects. On the contrary, high fidelity tactile sensing can be used for shape reconstruction of transparent objects as well as pose estimation and safe-manipulation~\cite{murali2021active, murali2022intelligent, kaboli2019tactile, kaboli2018active, kaboli2017tactile, kaboli2018robust, liu2022neuro, kaboli2016tactile}.

\begin{figure}[t!]
    \centering
    \includegraphics[width = 0.9\columnwidth]{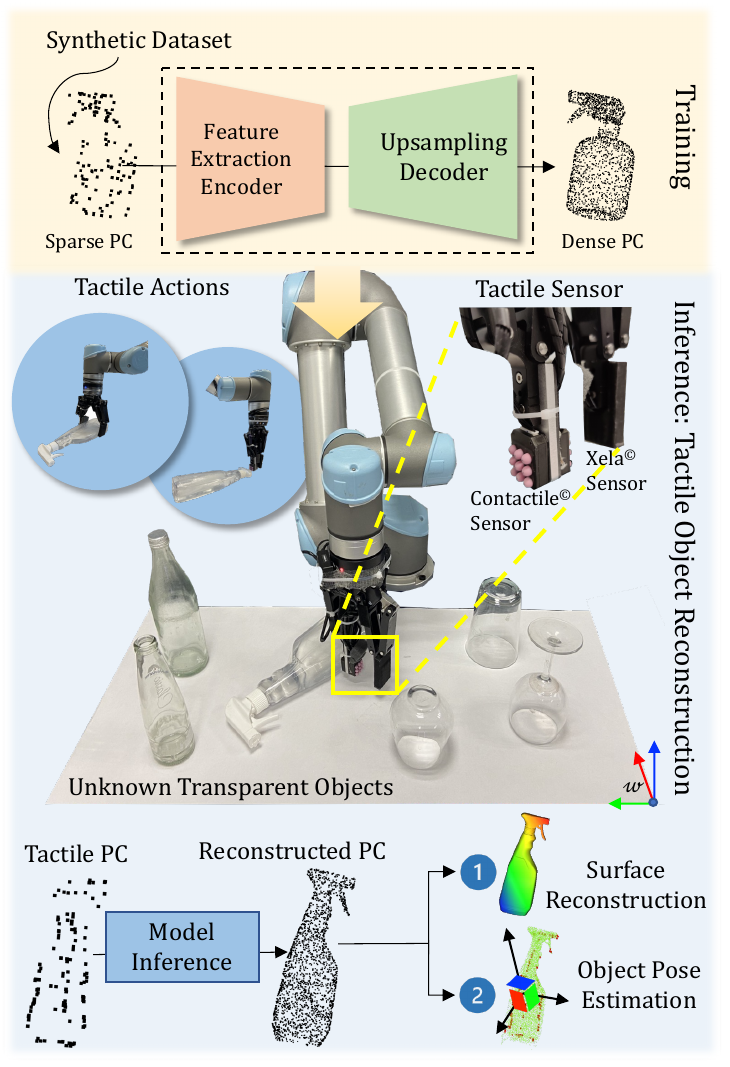}
    \caption{Experimental Setup: A Universal Robots UR5 with sensorised Robotiq Gripper with 3-axis tactile sensor arrays for active tactile-based category-level unknown transparent object reconstruction. }
    \label{fig:fig1}
\end{figure}

Tactile perception is inherently action-conditioned as data depends on the type of contact action performed and local as only the local surface information around the contact area is extracted~\cite{kaboli2015humanoids}. Hence, for reconstructing the surfaces of an object, multiple contact actions need to be performed by the robot. This leads to sparse information and prohibitively long data collection times.
Early works have used offline methods to collect dense tactile data and used shape-fitting primitives such as superquadratics~\cite{bierbaum2007haptic}. 
Aggregating contact points into a point cloud is often used to represent the shape of the objects. Some works have used Bayesian filtering techniques for defining a probabilistic model of the objects using the tactile point clouds and used them for other tasks such as classification~\cite{meier2011probabilistic}.
Gaussian process implicit surface (GPIS) has been widely used for tactile object reconstruction~\cite{dragiev2011gaussian, yi2016active, bjorkman2013enhancing, gandler2020object, martens2016geometric, suresh2021tactile, jamali2016active}. The implicit surface described by a Gaussian process describes the shape of an object through a function that decides for each point in space whether it is part of the object or not. It produces smooth surface manifolds with a reasonable number of tactile points as input and also provides probabilistic information to guide the tactile actions. However, for complex shapes it typically requires lots of points uniformly distributed on the object's surface for reconstruction~\cite{jamali2016active}. Some works have also used tactile sensing with visual perception in order to perform shape completion with prior information observed with visual cameras~\cite{gandler2020object, smith20203d}.  
While these works focus on opaque objects, limited works exist for the reconstruction of transparent objects. 
Recently, deep learning methods have been used for point cloud based shape completion given partial or noisy input point clouds~\cite{fei2022comprehensive, murali2022deep}. Seminal works on PointNet~\cite{qi2017pointnet} allowed using raw point clouds as inputs to deep networks for the task of classification and semantic segmentation. Prior works have worked towards point cloud completion using deep networks such as~\cite{yu2018pu, yuan2018pcn} but are mainly evaluated on datasets derived from CAD models and rarely evaluated on real-world platforms with noisy and sparse sensors~\cite{fei2022comprehensive}.     

Using the constructed object shape for pose estimation of a transparent object through tactile sensing brings further challenges due to the nature of the tactile data. Typical pose-estimation methods for visual perception perform poorly with tactile data as they are sparse and extracted sequentially through contact probing~\cite{Qiang-TRO-2020, piga2021maskukf, murali2021active, murali2022empirical, murali2022towards, murali2022deep}. 
In summary, there are limitations in the state-of-the-art for the reconstruction and further applications such as pose estimation of transparent objects with tactile perception: (a) existing reconstruction strategies such as GPIS fail to capture fine shape details with sparse tactile input data, (b) directly deploying deep learning based strategies for shape completion with sparse input data is impractical as the collection of a large dataset of tactile data for training is prohibitively expensive, (c) existing tactile-based pose estimation techniques rely upon known object models or shape primitives but category-level tactile-based pose estimation wherein objects without \textit{a priori} known CAD models but belong to a known category is necessary. 

\begin{figure*}[t!]
    \centering
    \includegraphics[width = \textwidth]{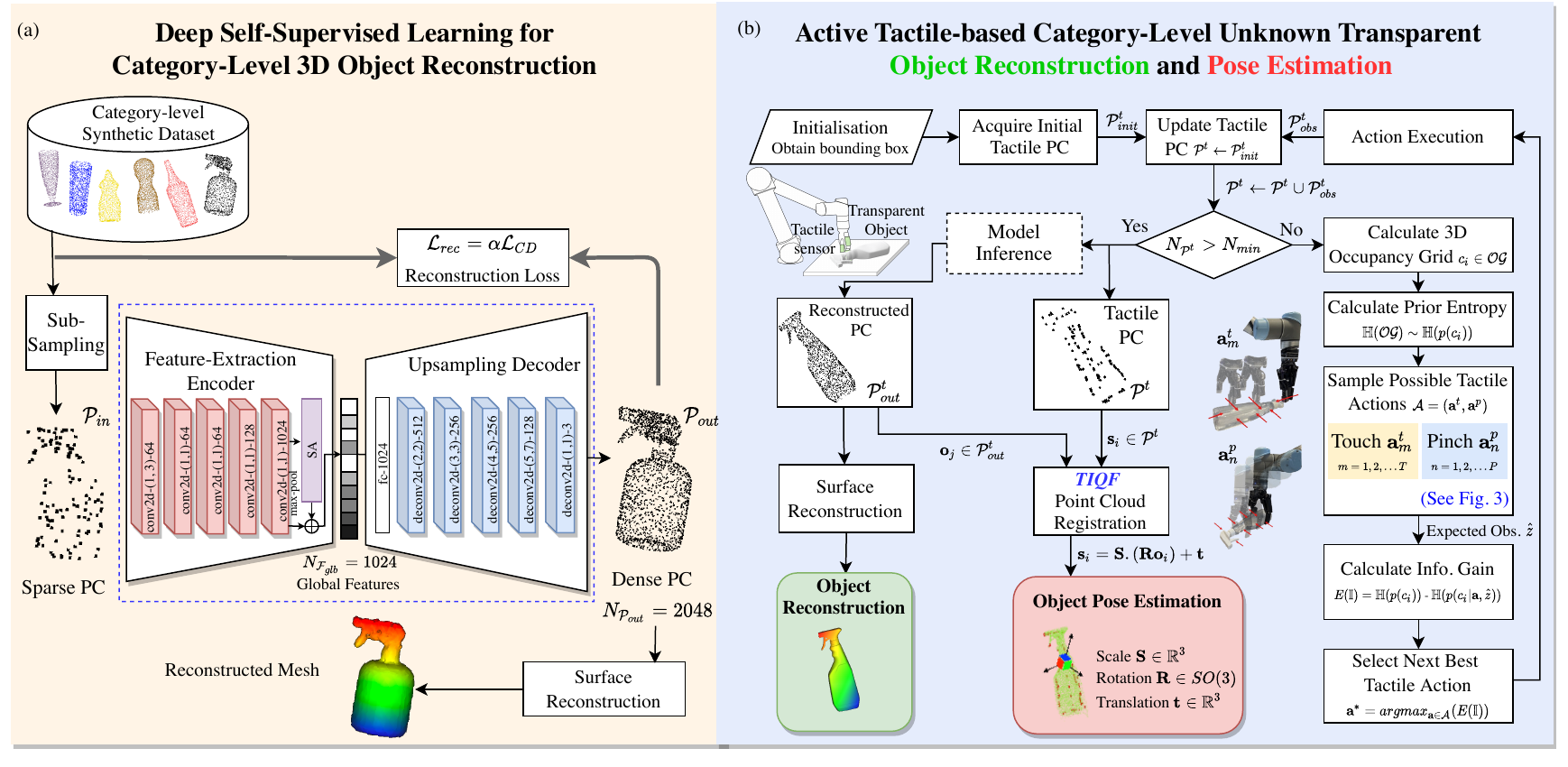}
    \caption{Proposed framework ACTOR: Active Tactile-based Category-Level Transparent Object Reconstruction.}
    \label{fig:framework}
\end{figure*}
\textbf{Contributions:} 
\begin{enumerate}[(I)]
    \item We propose \textit{ACTOR}, a novel framework for deep active tactile-based category-level perception of unknown transparent objects for reconstruction and pose estimation. Our proposed network is trained on a category-level synthetic dataset and tested on sparse tactile point clouds from real unknown transparent objects.
    \item Our proposed network consists of a feature-extraction encoder with self-attention and an upsampling decoder for accurate reconstruction of sparse input point clouds. 
    \item We propose an autonomous and active tactile-based unknown object exploration strategy based on information gain.
    \item We improve our previously presented novel Translation-Invariant Quaternion Filter (TIQF)~\cite{murali2022active} to category-level pose (6DoF) and scale (3DoF) estimation and relax the need of a prior known model of the object. 
\end{enumerate}
To validate our proposed framework, we perform extensive experiments on a real robotic setup and provide baseline comparisons with state-of-the-art methods for tactile-based object reconstruction and pose estimation.

\section{METHODS}
\label{sec:methods}
\subsection{Problem Definition and Proposed Framework}
The objective is to reconstruct a dense point cloud that precisely represents the shape of unknown transparent objects from sparse point clouds extracted with active tactile interactive perception. To this end, we propose a novel framework termed ACTOR shown in Fig.~\ref{fig:framework}. In Fig.~\ref{fig:framework}(a) we propose a self-surpervised learning approach with an autoencoder network that is trained on subsampled pointclouds from synthetic objects belonging to the same category but not identical as the real objects. In Fig.~\ref{fig:framework}(b), we propose a novel active tactile-based unknown transparent object exploration strategy which is used for inference with our trained model to reconstruct a dense point cloud. We demonstrate downstream tasks such as tactile-based pose estimation.

\subsection{Deep Self-Supervised Learning for 3D Object Reconstruction}
\label{ssec:deep_reconstruction}
We generate a dataset $\mathcal{D}$\footnote{\url{https://www.robotact.de/tactile-reconstruction}} of synthetic object models from the ShapeNet repository~\cite{chang2015shapenet} in order to leverage the open-source datasets and avoid expensive real tactile-data collection. The synthetic object models belong to the same category but are different from the real unknown transparent objects. 
We uniformly sample $N_{in} = 2048$ points from the synthetic object meshes. These pointclouds are normalized and scaled to fit into a $[0,1]^3$ cube and added to the dataset, $\mathcal{P}_{in} \in \mathcal{D}$. 
In order to generate the input point clouds $\mathcal{P}^{\bullet}_{in}$ to the network, we randomly subsample the $\mathcal{P}_{in}$ by voxel-grid subsampling by the factor $k$ i.e., $\mathcal{P}^{\bullet}_{in} \in \mathbb{R}^{\lceil \frac{1}{k}N_{in} \rceil \times 3}$.  This creates a challenging task for reconstruction with higher values for $k$ as simpler techniques based on interpolation with neighborhood points cannot be used. 

\subsubsection*{Feature-Extraction Encoder}
The network architecture shown in Figure~\ref{fig:framework}(a) is proposed as an autoencoder (AE) that uses a self-supervised approach to reconstruct the original point cloud from a subsampled point cloud. 
The encoder takes subsampled point clouds as inputs and generates a high dimensional feature vector. The feature vector captures the global geometric shape information of the input point cloud. 
In general, any deep network that works on raw input point clouds to provide a high dimensional feature vector can be used as an encoder. In particular,
we use a modified PointNet architecture~\cite{qi2017pointnet} for the encoder. PointNet takes unordered point clouds and generates a global feature descriptor vector of size 1024. The network learns a set of optimization functions that select interesting or informative points of the point cloud. The encoder consists of $[1\times1]$ convolutions with output channels size $(64, 64, 128, 1024)$ with the first convolutional layer with kernel size $[1\times3]$ to encode the input pointcloud of $N\times3$ dimension. The convolution layers are aggregated by a max-pooling layer. We introduce a self-attention layer~\cite{zhang2019self} whose outputs are aggregated with the max-pooled features to provide the global feature vector.  
We have summarized the encoder in Figure~\ref{fig:framework}(a).

\textbf{Self-Attention (SA) Layer:} The SA layer is introduced as it can encode meaningful spatial relationships between features and focus on important local features. From the input layer ($\mathtt{conv2d-1024}$), two separate multi-layer perceptrons (MLPs) are used to get features $\mathbf{G}$ and $\mathbf{H}$ which are subsequently used to get the weights as $\mathbf{W} = softmax(\mathbf{G}^T\mathbf{H})$. The input features are transformed using another MLP to obtain $\mathbf{K}$ and multiplied with the weights as $\mathbf{W}^T\mathbf{K}$.
These vectors are summed with the input vector to produce the output features.

\subsubsection*{Upsampling Decoder}
We design an upsampling decoder that upsamples the input global feature vector to provide the reconstructed dense output point cloud $\mathcal{P}_{out}$. The upsampling decoder is composed by a fully connected layer with output dimension of 1024 and five deconvolutional layers with kernel sizes and output channels shown in Fig.~\ref{fig:framework}(a).  
The decoder produces the output point cloud with point size set to 2048 while training as this is sufficiently dense for reconstruction purposes. 

\subsubsection*{Loss Function}
In order to encourage the upsampled point cloud to be in proximity to the original input point cloud and follow the underlying geometrical surface of the object, we use the Chamfer distance metric~\cite{borgefors1986distance} as the loss. Given the input point cloud prior to subsampling, $\mathcal{P}_{in}$ and the reconstructed output point cloud $\mathcal{P}_{out}$, the loss is defined as:
\begin{align}
    \mathcal{L}_{CD}(\mathcal{P}_{in}, \mathcal{P}_{out}) &= \frac{1}{|\mathcal{P}_{in}|}\sum_{p_1 \in \mathcal{P}_{in}} \min_{p_2 \in \mathcal{P}_{out}} ||p_1 - p_2||_{2} + \\ & \frac{1}{|\mathcal{P}_{out}|}\sum_{p_2 \in \mathcal{P}_{out}} \min_{p_1 \in \mathcal{P}_{in}} ||p_2 - p_1||_{2} \nonumber,
    \label{eq:chamfer_dist}
\end{align}
where $|\bullet|$ refers to the number of points in the point cloud and $||\bullet||_2$ refers to the L2 norm. The loss $\mathcal{L}_{CD}$ represents the average distance between the \textit{closest} points in the two point clouds. We use the weighted loss for learning stability as the reconstruction loss $\mathcal{L}_{rec} = \alpha\mathcal{L}_{CD}$ with $\alpha = 100$ set empirically.
For surface reconstruction from the dense reconstructed point cloud, we use the ball-pivoting algorithm~\cite{bernardini1999ball}.


\subsection{Active Deep Tactile-based Unknown Transparent Object Reconstruction and Pose Estimation}
\subsubsection{Active Tactile-based Transparent Object Reconstruction}
\begin{figure}[t!]
    \centering
    \includegraphics[width = 0.9\columnwidth]{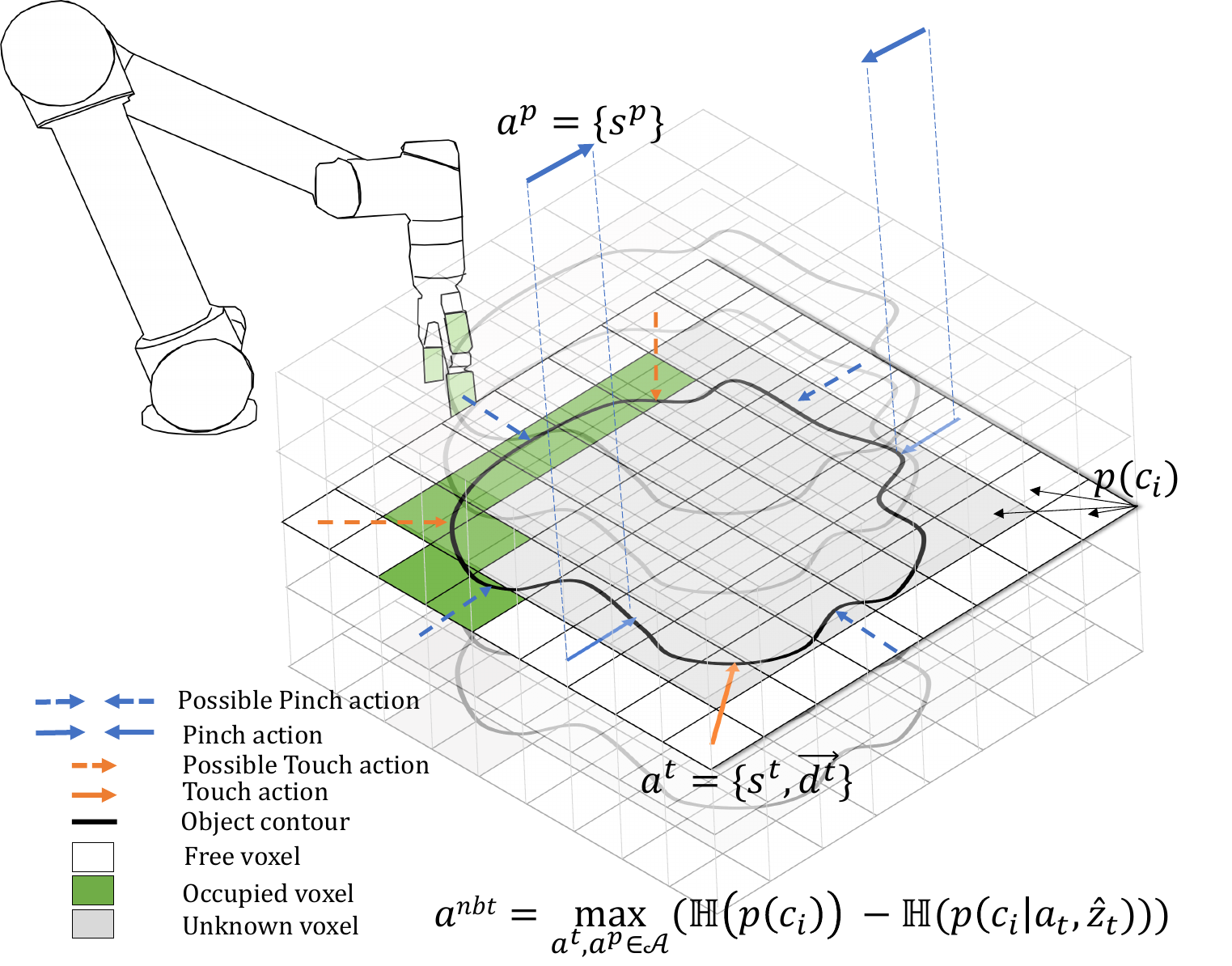}
    \caption{Action selection voxelised probabilistic occupancy grid.}
    \label{fig:occupancy_grid}
\end{figure}

The model trained with only \textit{synthetic data} as described in Sec.~\ref{ssec:deep_reconstruction} is used during the inference with \textit{real-world} transparent objects. The sparse tactile point cloud data is collected autonomously by the robot using an information gain-based active strategy. We define two types of tactile actions for data acquisition: touch and pinch actions as shown in Figure~\ref{fig:occupancy_grid}.
The touch action is executed as a guarded horizontal straight-line motion wherein the object is not moved upon contact. The touch action is defined by a tuple $\mathbf{a}^{t} = \{\mathbf{s}^t, \overrightarrow{\mathbf{d}^t} \}$ where $\mathbf{s}^t \in \mathbb{R}^3$ is the start point of the tactile-sensorised gripper and $\overrightarrow{\mathbf{d}^t} \in \mathbb{R}^3$ is the direction of the gripper-motion defined in the world-coordinate frame $\mathcal{W}$. During the pinch action the robot approaches the object in a vertical straight-line motion with a completely open gripper and performs an antipodal enclosure grasp on the object. The fingers of the gripper are closed until the force on the tactile sensors exceeds a predefined threshold.
The pinch action is characterized by $\mathbf{a}^{p} = \{\mathbf{s}^p \}$ where $\mathbf{s}^p \in \mathbb{R}^3 $ is the start position of the gripper motion vertically above the object at a predefined height as shown in Figure~\ref{fig:occupancy_grid}. Given the 2D bounding box of the object (a priori known or through a RGB camera), a probabilistic occupancy grid $\mathcal{OG}_i$ of preset height and resolution $og_{res}$ is defined. Each cell of the occupancy grid $c_i$ is represented by an occupancy probability $p(c_i)$ which is initially set to 0.5. During exploration, if a cell is discovered to belong to the object, the probability is set to 1 and similarly, if the cell belongs to free space, the probability is set to 0. The probabilities are updated through ray intersections based on the virtual sensor model. We define a virtual sensor model of the tactile sensor which casts a set of rays $\mathcal{R} = \{r_1, r_2, \dots, r_{n_{taxel}} \}$ where ${n_{taxel}} $ refers to the number of taxels in the sensor array. The independence assumption of the probability of each grid cell with one another allows us to calculate the overall entropy of the $\mathcal{OG}$ as the summation of the entropy of each cell. The Shannon entropy of the overall occupancy grid is calculated as:
\begin{equation}
    \mathbb{H}(\mathcal{OG}) = \sum_{c_i \in \mathcal{OG}} p(c_i)log(p(c_i)) + (1 - p(c_i))(1 - log(p(c_i))).
    \label{eq:entropy}
\end{equation}
Monte-Carlo sampling of possible tactile actions $N_{nbt}$ are performed for computing the next best tactile (NBT) action. The actions space $\mathcal{A}_{nbt}$ is comprised of an equal number of touch and pinch respectively as $\mathcal{A}_{nbt} = \{a^p, a^t\}_{N_{nbt}}$. The expected measurements $\hat{\mathbf{z}}_t$ for each action $a_t \in \mathcal{A}$ is computed using ray-traversal algorithms~\cite{hornung2013octomap}. 
Given the observed grid cell $c$ and the measurement from sensor observation $z$, the log-odds is updated as $L(c|z) = L(c) + l(z)$ wherein $L(c) = log\frac{p(c)}{1-p(c)}$ and  
\begin{equation}
    l(z) = \left\{
                \begin{array}{ll}
                  log\frac{p_h}{1-p_h}  \quad \mathrm{if} \ z \widehat{=} \textit{ hit} \\
                  log\frac{p_m}{1-p_m} \quad \mathrm{if} \ z \widehat{=} \textit{ miss} 
                \end{array}
              \right.
    \label{eq:log-odds}
\end{equation}
where $p_h$ and $p_m$ are the probabilities of hit and miss which are user-defined values set to 0.7 and 0.4 respectively as in~\cite{hornung2013octomap}. The posterior probability $p(c|z)$ can be computed by inverting $L(c|z)$. The expected information gain by taking an action $a_t \in \mathcal{A}_{nbt}$ with expected measurement $\hat{\mathbf{z}}_t$ is provided by the Kullback-Liebler divergence of the posterior entropy and the prior entropy as: 
\begin{equation}
    E[\mathbb{I}(p(c_i | \mathbf{a}_t,  \hat{z}_t))] = \mathbb{H}(p(c_i)) - \mathbb{H}(p(c_i | \mathbf{a}_t,  \hat{z}_t))
    \label{eq:kl_view}
\end{equation}
Therefore, the action that maximizes the expected information gain is considered as the NBT action:
\begin{equation}
    \mathbf{a}^{nbt*}_t = \argmax_{\mathbf{a} \in \mathcal{A}}(E[\mathbb{I}(p(c_i | \mathbf{a}_t,  \hat{z}_t))])
    \label{eq:kl_view_max}
\end{equation}
Each tactile action extracts contact positions in 3D space and contact forces. The direction of the normal force is used to extract the normal direction $\hat{n}$ of the object surface. The contact points are aggregated into the tactile point cloud $\mathcal{P}^t$. In order to initialize the NBT action calculation, an initial point cloud (with $N_{\mathcal{P}^t} = 20$) is required, which is extracted by randomised touch actions. Further points are collected in an active manner using the NBT criteria. A minimum number of points in the tactile point cloud is required to perform model inference $N_{\mathcal{P}^t} > N_{min}$ which is tuned empirically. The tactile point cloud is provided as input to the trained network and the reconstructed point cloud $\mathcal{P}_{out}$ is obtained . 


\subsubsection{Tactile-Based Object Pose Estimation}
\label{ssec:pose_estimation}

We perform the 6D pose estimation through dense to sparse point cloud registration. The sparse scene point cloud $\mathbf{s}_i \in \mathcal{S}$ is represented by the tactile points and the dense object point cloud $\mathbf{o}_i \in \mathcal{O}$ is represented by the reconstructed point cloud in~\ref{ssec:deep_reconstruction} without the need for the object model. Point cloud registration problem with $M$ known correspondences can be formulated as:
\begin{equation}
     \mathbf{s}_i = \mathbf{S}\cdot(\mathbf{R}\mathbf{o}_i) + \mathbf{t} \quad i = 1, \dots M,
     \label{eq:generativemodel}
 \end{equation}
where $\mathbf{S} \in \mathbb{R}^3$ represents scale, $\mathbf{R} \in SO(3)$ represents rotation and $\mathbf{t} \in \mathbb{R}^3$ represents translation which are unknown and to be estimated and $\cdot$ is the element-wise product. 

We perform the point cloud registration using our novel translation-invariant Quaternion filter (TIQF) presented in~\cite{murali2022active} to determine $\mathbf{R}$, $\mathbf{S}$ and $\mathbf{t}$. 
The scale, rotation and translation are decoupled by finding the relative vectors between corresponding points, i.e., $\forall o_i, o_j \in \mathcal{O}, s_i, s_j \in \mathcal{S}$ the relative vectors are $\mathbf{s}_{ji} = \mathbf{s}_j - \mathbf{s}_i$ and $\mathbf{o}_{ji} = \mathbf{o}_j - \mathbf{o}_i$. Equation~\eqref{eq:generativemodel} is reformulated as:
\begin{align}
    \mathbf{s}_j - \mathbf{s}_i &= (\mathbf{S}\cdot\mathbf{R}\mathbf{o}_j + \mathbf{t}) - (\mathbf{S}\cdot\mathbf{R}\mathbf{o}_i + \mathbf{t}) ,\\
    \mathbf{s}_{ji} &= \mathbf{S}\cdot\mathbf{R}\mathbf{o}_{ji} \quad .
    \label{eq:trans_invariance}
\end{align}

We note that equation~\eqref{eq:trans_invariance} is independent of translation. Taking the L2-norm on both sides for Eq.~\eqref{eq:trans_invariance} and recalling that norm is rotation invariant we get:
\begin{equation}
    \mathbf{||s||}_{ji} = \mathbf{||S||}\cdot\mathbf{||o||}_{ji} \quad .
    \label{eq:rot_invariance}
\end{equation}
The scale $\mathbf{S}$ is estimated by taking the ratio of the axis aligned bounding box (AABB) of the scene and object point clouds, i.e., if $\mathcal{X}_{AABB} = \{ (x_{min}, x_{max}), (y_{min}, y_{max}), (z_{min}, z_{max}) \}$ represents the AABB for a point cloud $\mathcal{X}$, then:
\begin{align}
     \mathbf{S} &= \{ \frac{|x_{max} - x_{min}|_{\mathcal{S}}}{|x_{max} - x_{min}|_{\mathcal{O}}}, \frac{|y_{max} - y_{min}|_{\mathcal{S}}}{|y_{max} - y_{min}|_{\mathcal{O}}} , \frac{|z_{max} - z_{min}|_{\mathcal{S}}}{|z_{max} - z_{min}|_{\mathcal{O}}}    \}
     \label{eq:scale}
 \end{align}
Using the estimated scale and using $\tilde{\mathbf{o}}_{ji} = \mathbf{S}\mathbf{o}_{ji}$ for convenience we are left with a pure rotation to estimate:  
\begin{align}
    \tilde{\mathbf{s}}_{ji} &= \mathbf{R}\tilde{\mathbf{o}}_{ji} \quad .
    \label{eq:trans_scale_invariance}
\end{align}
 We cast the rotation estimation problem into a recursive Bayesian estimation framework and derive a linear state and measurement model. Reformulating Eq.\eqref{eq:trans_scale_invariance} using quaternions we get: 
 \begin{equation}
    \overline{\mathbf{s}}_{ji} = \mathbf{x} \odot \overline{\mathbf{o}}_{ji} \odot \mathbf{x}^{*}, 
    \label{eq:quat_objective}
\end{equation}
where $\mathbf{x}$ is the quaternion form of $\mathbf{R}$, $\odot$ is the quaternion product, ${\mathbf{x}}^{*}$ is the conjugate of $\mathbf{x}$, and $\overline{\mathbf{s}}_{ji}=\{0,\tilde{\mathbf{s}}_{ji}\}$ and $\overline{\mathbf{o}}_{ji}=\{0,\tilde{\mathbf{o}}_{ji}\}$.
Using the matrix form of quaternion product, we can rewrite Eq.\eqref{eq:quat_objective} as:
\begin{align}
    \begin{bmatrix}
        0 & -\tilde{\mathbf{s}}_{ji}^T \\
        \tilde{\mathbf{s}}_{ji} & \tilde{\mathbf{s}}_{ji}^{\times}
    \end{bmatrix}\mathbf{x} -  \begin{bmatrix}
        0 & -\tilde{\mathbf{o}}_{ji}^T \\
        \tilde{\mathbf{o}}_{ji} & -\tilde{\mathbf{o}}_{ji}^{\times}
    \end{bmatrix} \mathbf{x} = \mathbf{0} \\
    \underbrace{\begin{bmatrix}
        0 & -(\tilde{\mathbf{s}}_{ji} - \tilde{\mathbf{o}}_{ij})^T \\
        (\tilde{\mathbf{s}}_{ji} - \tilde{\mathbf{o}}_{ji}) & (\tilde{\mathbf{s}}_j + \tilde{\mathbf{s}}_i + \tilde{\mathbf{o}}_j + \tilde{\mathbf{o}}_i)^{\times}
        \end{bmatrix}_{4 \times 4}}_{\mathbf{H}_t} \mathbf{x} &= \mathbf{0} \quad ,
        \label{eq:expected_measurement}
\end{align}
where $(\ )^\times$ denotes the skew-symmetric matrix formulation. Equation~\eqref{eq:expected_measurement} is of the form $\mathbf{H}_t\mathbf{x} = 0$ where $\mathbf{H}_t$ is the pseudo-measurement matrix~\cite{choukroun2006novel}. We note that Eq.~\eqref{eq:expected_measurement} represents a noise-free state estimation where $\mathbf{H}_t$ depends only on sparse and dense point correspondences which are $\tilde{\mathbf{s}}_{ji}$ and $\tilde{\mathbf{o}}_{ji}$. We design a pseudo-measurement model as $ \mathbf{H}_t \mathbf{x} = \mathbf{z}^h$
and set $\mathbf{z}^h = 0$. Since we have a static process model, the object does not move and $\mathbf{x}$ and $\mathbf{z}_t$ are Gaussian distributed, 
the state $\mathbf{x}_t$ and covariance matrix $\Sigma^{\mathbf{x}}_{t}$ at each timestep $t$ are computed through a linear Kalman filter. The Kalman filter equations are skipped for brevity and a in-depth derivation is provided in our prior work~\cite{murali2022active}.
As the Kalman filter does not implicitly ensure the constraints on the quaternion as $||\mathbf{x}|| = 1$, we normalise the state and uncertainty after each update step as $\bar{\mathbf{x}}_{t} = \frac{\mathbf{x}_{t}}{||\mathbf{x}_{t}||_2} \quad, \bar{\Sigma}^{\mathbf{x}}_{t} = \frac{\Sigma^{\mathbf{x}}_{t}}{||\mathbf{x}_{t}||_2^2}$. We convert the estimated rotation $\Bar{\mathbf{x}}_t$ to its equivalent rotation matrix $\mathbf{R}$. It used to estimate the translation using the following relation: $\mathbf{t} = \frac{1}{N} \sum_{i=0}^{N} (\Bar{\mathbf{s}}_i - \mathbf{R} \Bar{\mathbf{o}}_i).$
At each iteration, a rotation and translation estimate is found which is used to transform the object point cloud and the process is repeated by re-estimating the correspondence points. The convergence criteria are set by (a) maximum number of iterations or (b) the relative change in estimated pose parameters is less than a predefined threshold ($0.1mm$ and $0.1^o$). 

\section{EXPERIMENTAL RESULTS}
\label{sec:experiment}
\subsection{Experimental Setup}
\label{ssec:setup}
The experimental setup is shown in Fig.~\ref{fig:fig1} consists of a set of 9 unknown transparent objects belonging to six categories and a Universal Robots UR5 equipped with a sensorised Robotiq 2F140 gripper. The tactile sensor array of the two-finger gripper are sourced from XELA robotics\textsuperscript\textcopyright and Contactile\textsuperscript\textcopyright. The outer and inner side of each finger are sensorised and comprise of $3\times3$ sensor array from the Contactile sensors and $4 \times 4$ sensor array from XELA sensors respectively. The fingertip of the finger sensorised with the XELA sensors also has $6 \times 1$ array. Each taxel of the sensor array provides 3-axis force measurements.
The normalised force values of the tactile sensors are measured and contact is established when the force exceeds the baseline threshold $f_{ts} \geq \tau_f$ where $\tau_f = 1.1$.
All operations involving point clouds use the Point Cloud Library\footnote{\url{https://pointclouds.org/}}, occupancy grid computations uses Octomap library\footnote{\url{https://octomap.github.io/}}, and the overall setup uses a ROS-based framework\footnote{\url{https://www.ros.org/}}. All robot experiments are run on a workstation using Ubuntu 18.04 with Intel\textsuperscript\textcopyright Xeon(R) Gold 5222 CPU. The object exploration and reconstruction time is between 5-7 minutes on average as the robot's maximum speed is limited to 250 mm/s for safety regulations.
\subsubsection*{Network Implementation Details} Our proposed network is implemented using the Tensorflow framework and training/ inference are performed on Nvidia Quadro RTX 4000 GPU. We used the ADAM optimiser, learning rate set to $10^{-4}$, momentum 0.9 and batch size 8. All layers of the encoder-decoder uses batch normalisation and the decay rate initialized at 0.5 and gradually increased to 0.99 with decay step size $2\times 10^5$. During training with our synthetic dataset $\mathcal{D}$, random voxel-grid subsampling is done to have input point clouds with point size between 40 and 120. 
\subsubsection*{Object List} We use the following widely-available transparent objects as unknown objects: bottle 1, bottle 2, can, detergent, cup 1, cup 2, cup 3, wineglass and spray as shown in Tab.~\ref{tab:qualitative_results}.

\subsection{Active Tactile-based Deep Self-Supervised Category-level Transparent Object Reconstruction}
\begin{figure*}[t!]
    \centering
    \begin{subfigure}[b]{0.32\textwidth}
    \includegraphics[width=\textwidth]{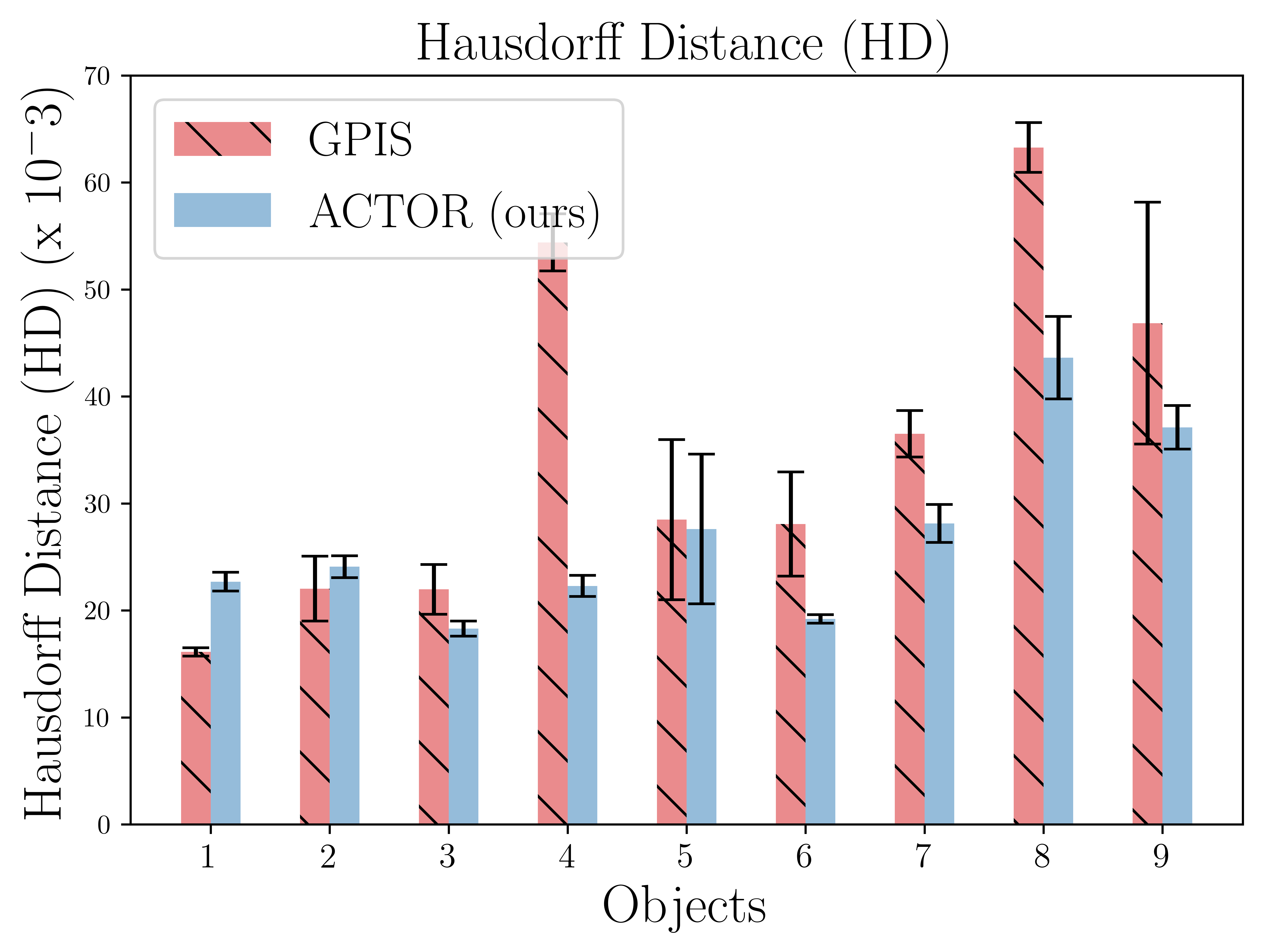}
    \caption{Hausdorff distance (HD)}
    \label{fig:hd_plot}
    \end{subfigure}
    \begin{subfigure}[b]{0.32\textwidth}
    \includegraphics[width=\textwidth]{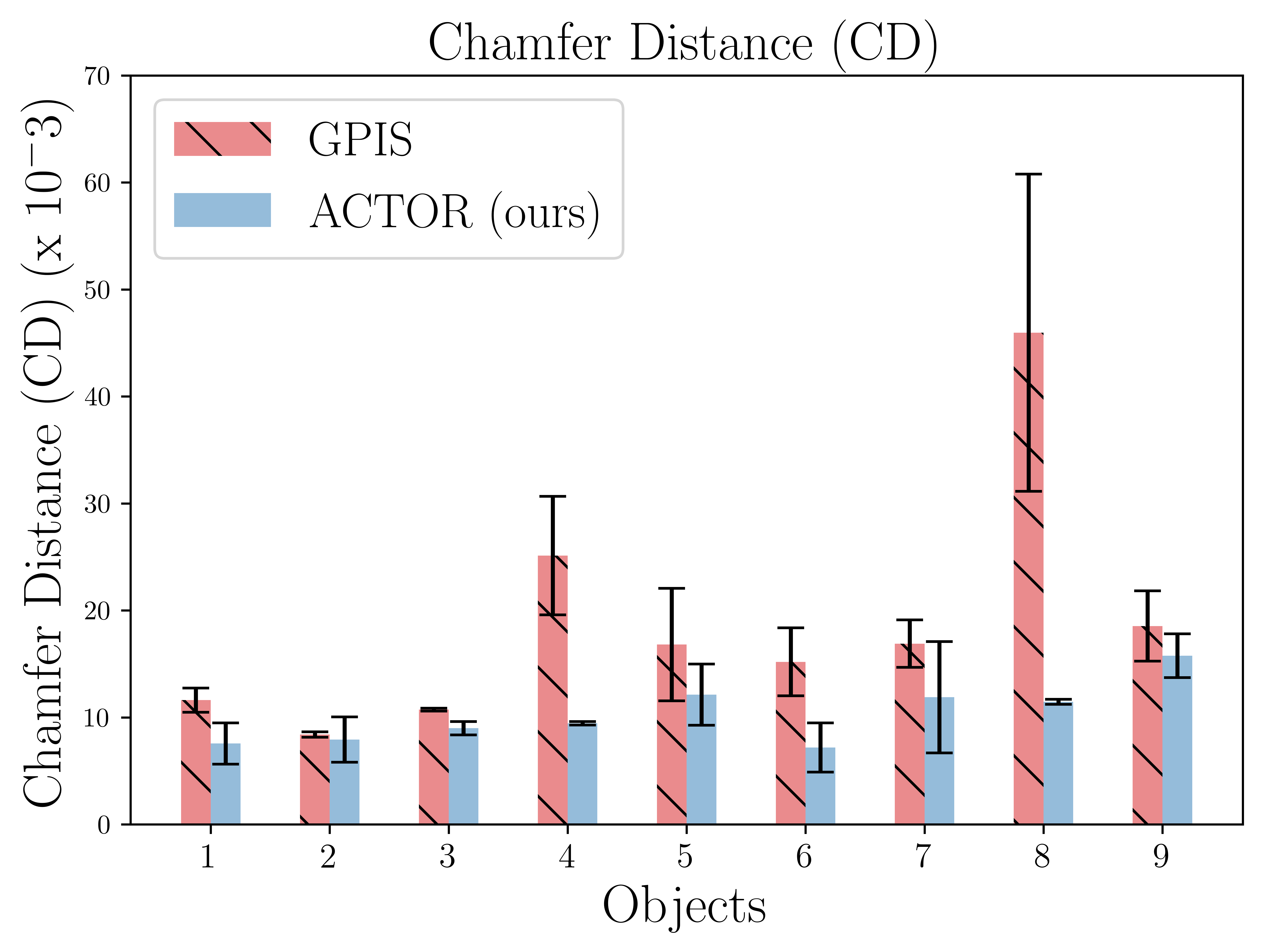}
    \caption{Chamfer distance (CD)}
    \label{fig:cd_plot}
    \end{subfigure}
    \begin{subfigure}[b]{0.32\textwidth}
    \includegraphics[width=\textwidth]{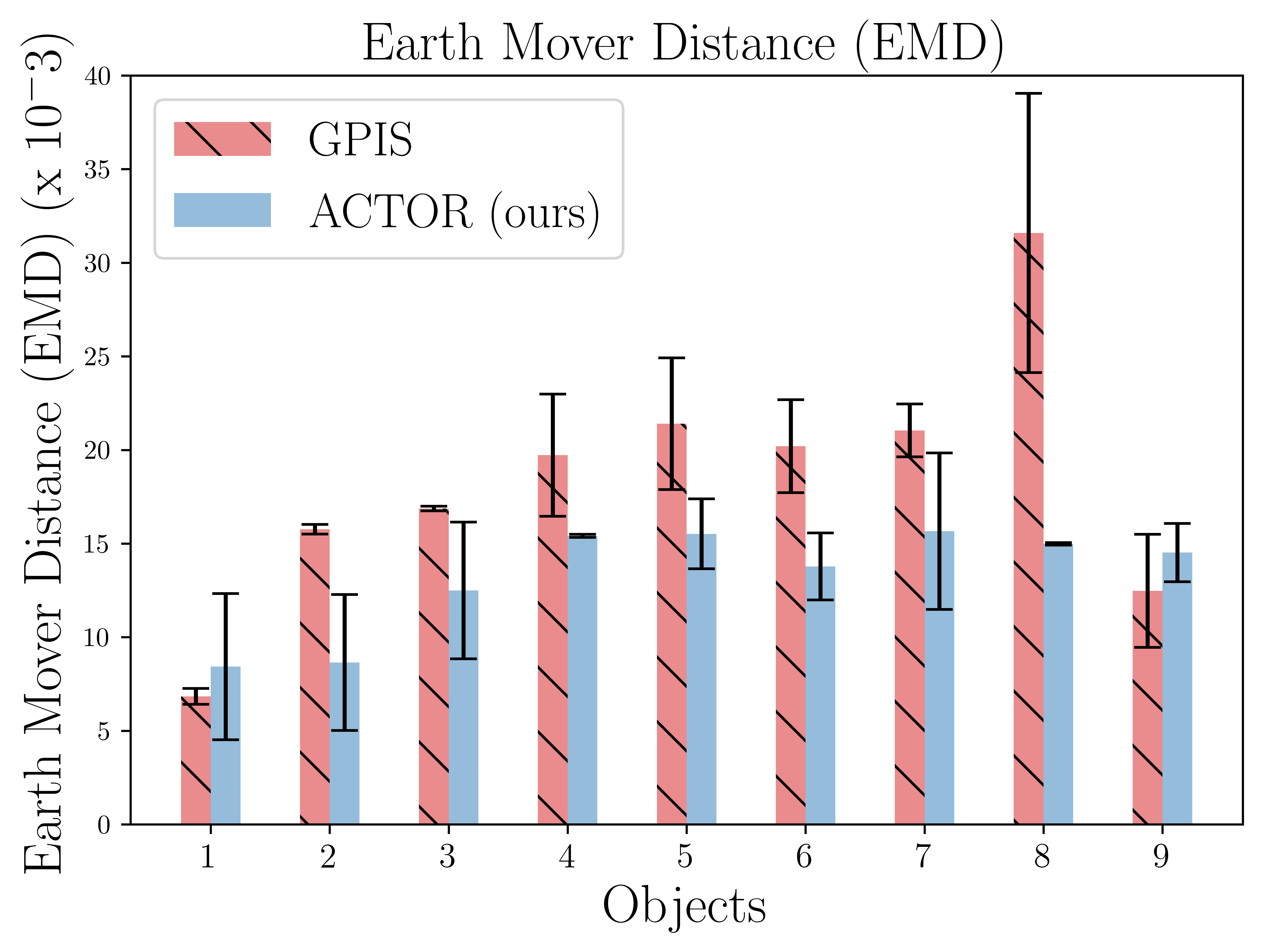}
    \caption{Earth mover distance (EMD)}
    \label{fig:emd_plot}
    \end{subfigure}
    \caption{Quantitative reconstruction results. Object numbered as follows: \{1: Bottle 1, 2: Bottle 2, 3: Can, 4: Detergent, 5: Cup 1, 6: Cup 2, 7: Cup 3, 8: Wineglass, 9: Spray \}}
    \label{fig:quant_plots}
\end{figure*}
\begin{table*}[t!]
\centering
\caption{Qualitative reconstruction results of our proposed method in comparison with Gaussian process implicit surfaces for unknown real test objects. (Best viewed on screen in color).}
\label{tab:qualitative_results}
\resizebox{\textwidth}{!}{%
\begin{tabular}{@{}ll|l|cc|cl|cc@{}}
\toprule
\multicolumn{2}{c}{Object}   & \multicolumn{1}{c}{Tactile PC}  & \multicolumn{2}{c}{Ground Truth} & \multicolumn{2}{c}{GPIS}   & \multicolumn{2}{c}{ACTOR (ours)} \\
         & & N $\sim$120  & PC           & Surface          & Recon. PC & Recon. Surf. & Recon. PC    & Recon. Surf.   \\ \midrule
Bottle 1 & \parbox[c]{1em}{
\includegraphics[height=1cm]{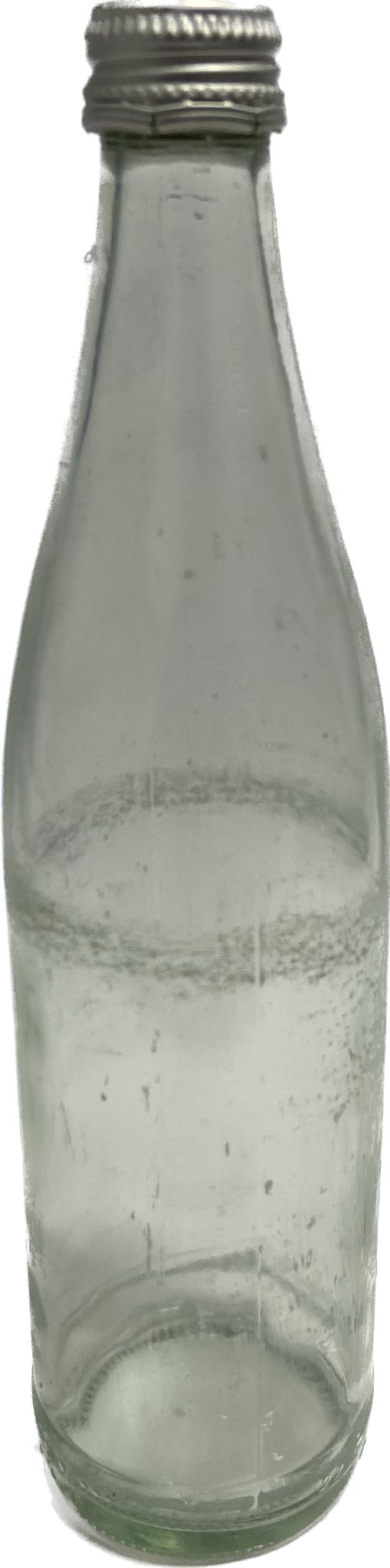}}    &   \parbox[c]{1em}{
\includegraphics[height=1cm]{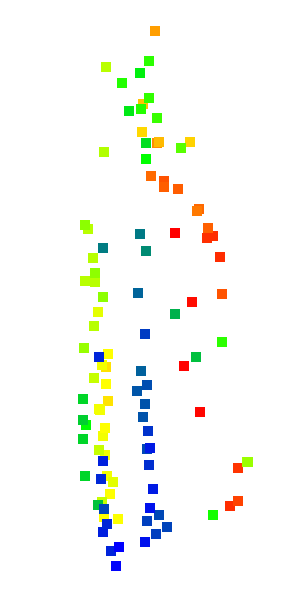}}        & \parbox[c]{1em}{\includegraphics[height=1cm]{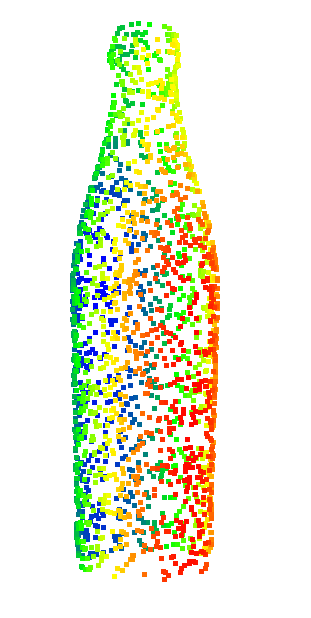}}            &    \parbox[c]{1em}{\includegraphics[height=1cm]{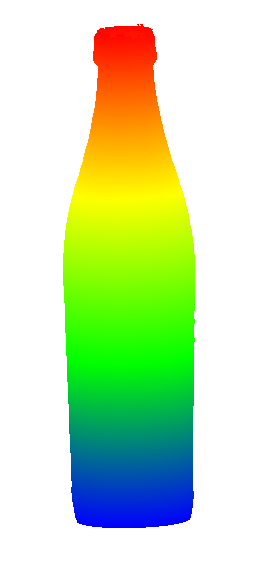}}        & 
\parbox[c]{1em}{\includegraphics[height=1cm]{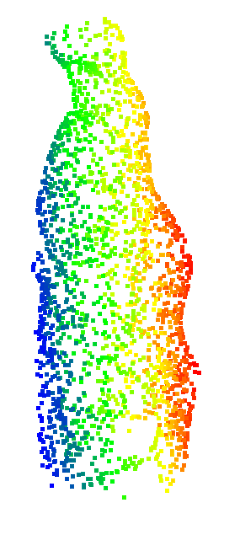}}           & 
\parbox[c]{1em}{\includegraphics[height=1cm]{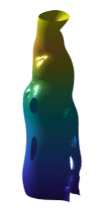}}              &
\parbox[c]{1em}{\includegraphics[height=1cm]{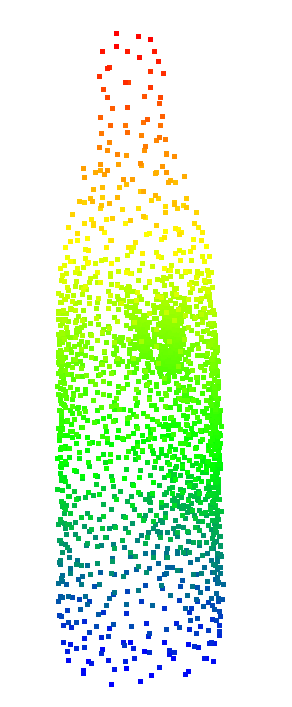}}
& \parbox[c]{1em}{\includegraphics[height=1cm]{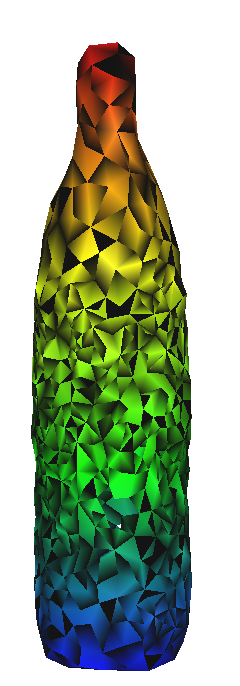}}  
             \\
Bottle 2 &\parbox[c]{1em}{
\includegraphics[height=1cm]{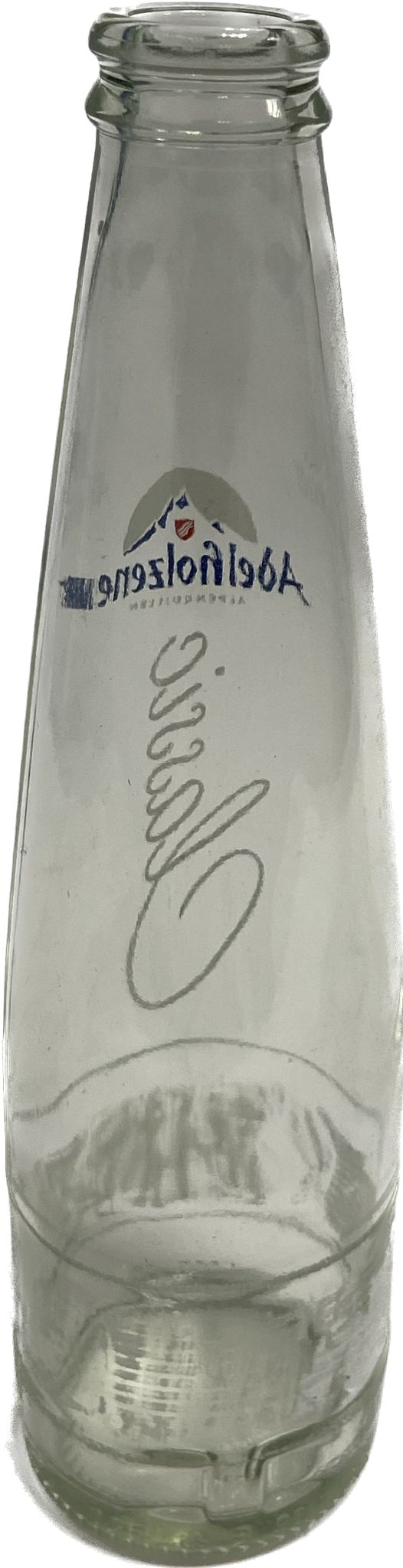}} &   \parbox[c]{1em}{
\includegraphics[height=1cm]{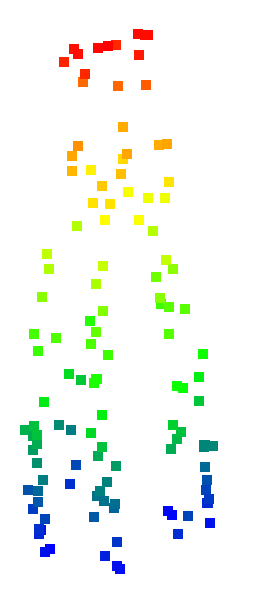}}      & \parbox[c]{1em}{\includegraphics[height=1cm]{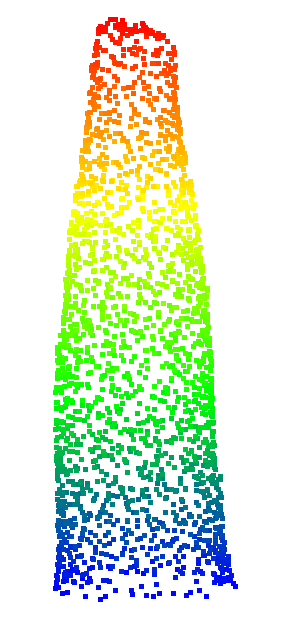}}              &   \parbox[c]{1em}{\includegraphics[height=1cm]{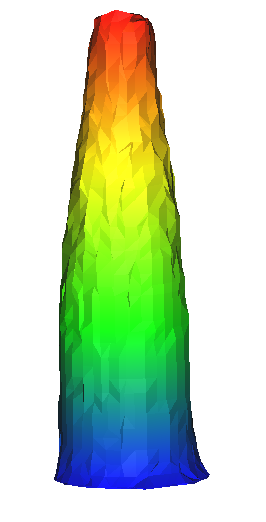}}                &    \parbox[c]{1em}{\includegraphics[height=1cm]{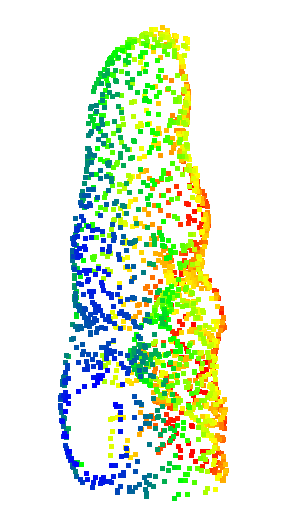}}       &    \parbox[c]{1em}{\includegraphics[height=1cm]{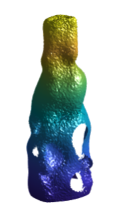}}            &   \parbox[c]{1em}{\includegraphics[height=1cm]{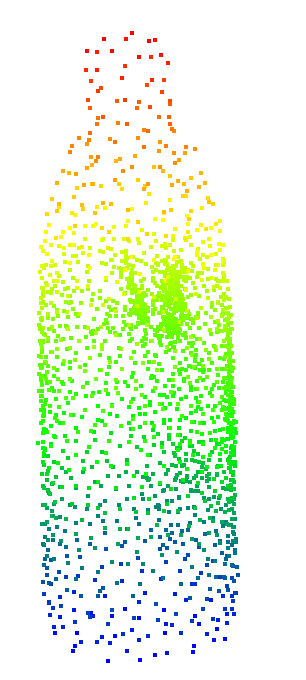}}           &   \parbox[c]{1em}{\includegraphics[height=1cm]{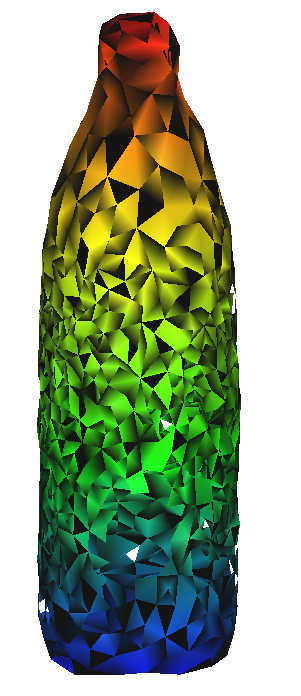}}                         \\
Can     &\parbox[c]{1em}{
\includegraphics[height=1cm]{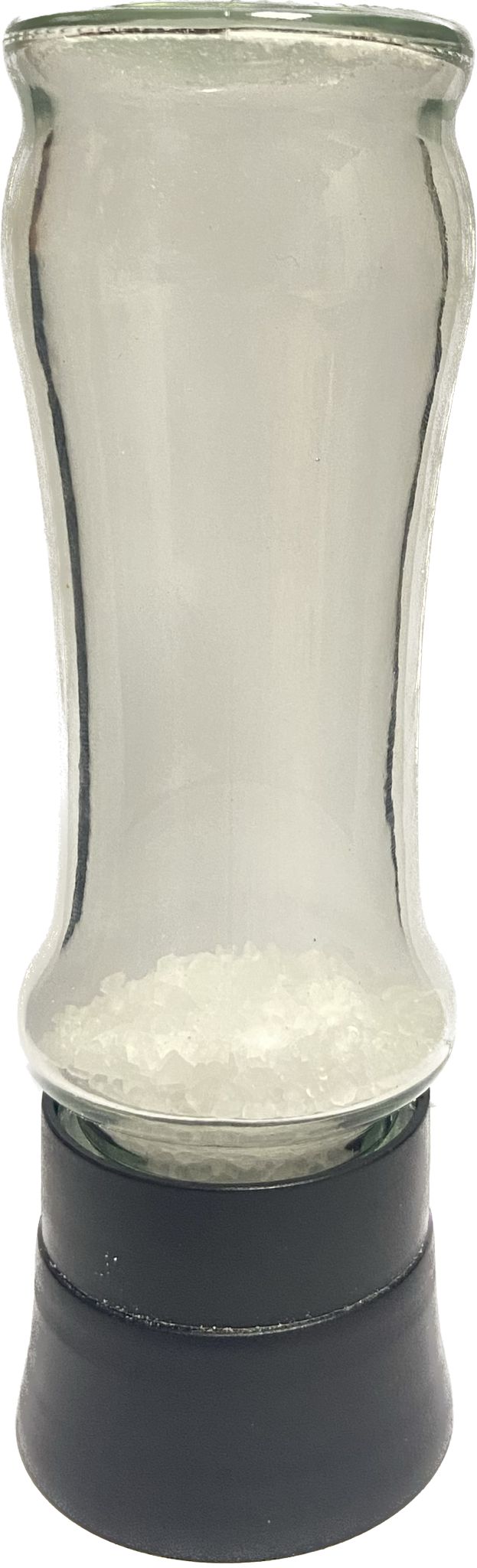}}   &     \parbox[c]{1em}{
\includegraphics[height=1cm]{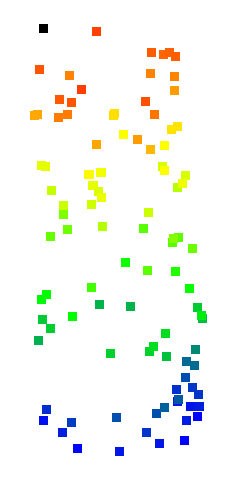}}        &      \parbox[c]{1em}{\includegraphics[height=1cm]{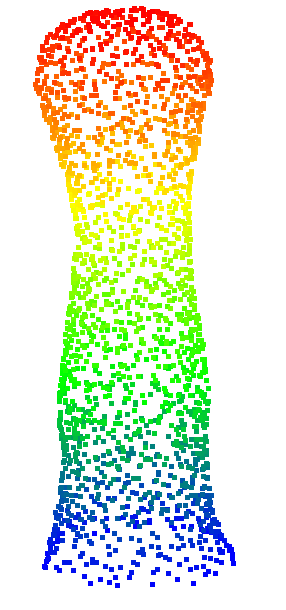}}         &     \parbox[c]{1em}{\includegraphics[height=1cm]{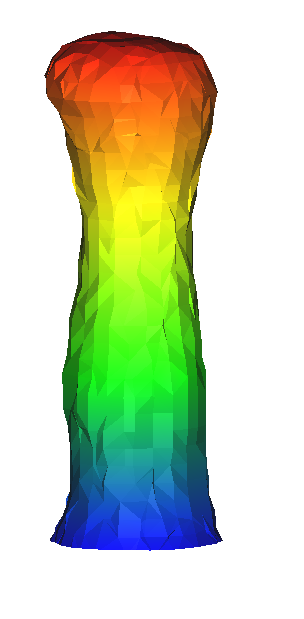}}        &  \parbox[c]{1em}{\includegraphics[height=1cm]{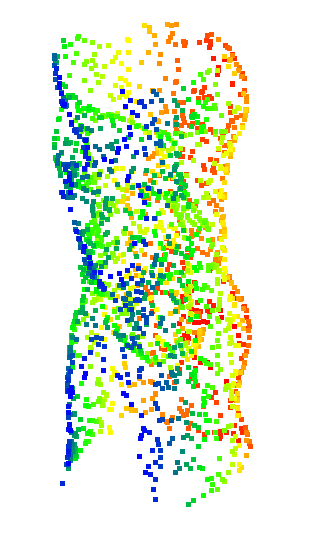}}         &     \parbox[c]{1em}{\includegraphics[height=1cm]{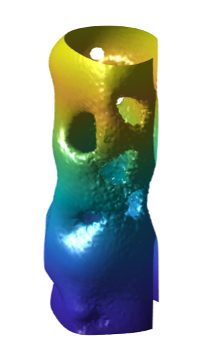}}           &    \parbox[c]{1em}{\includegraphics[height=1cm]{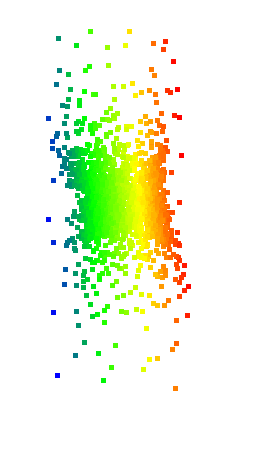}}          &     \parbox[c]{1em}{\includegraphics[height=1cm]{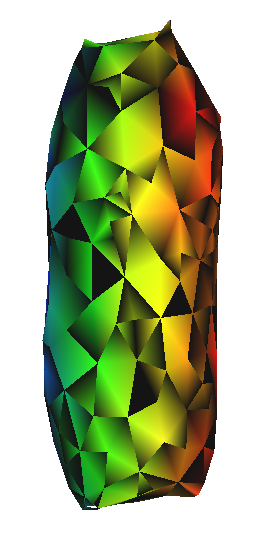}}                       \\
Detergent &\parbox[c]{1em}{
\includegraphics[height=1cm]{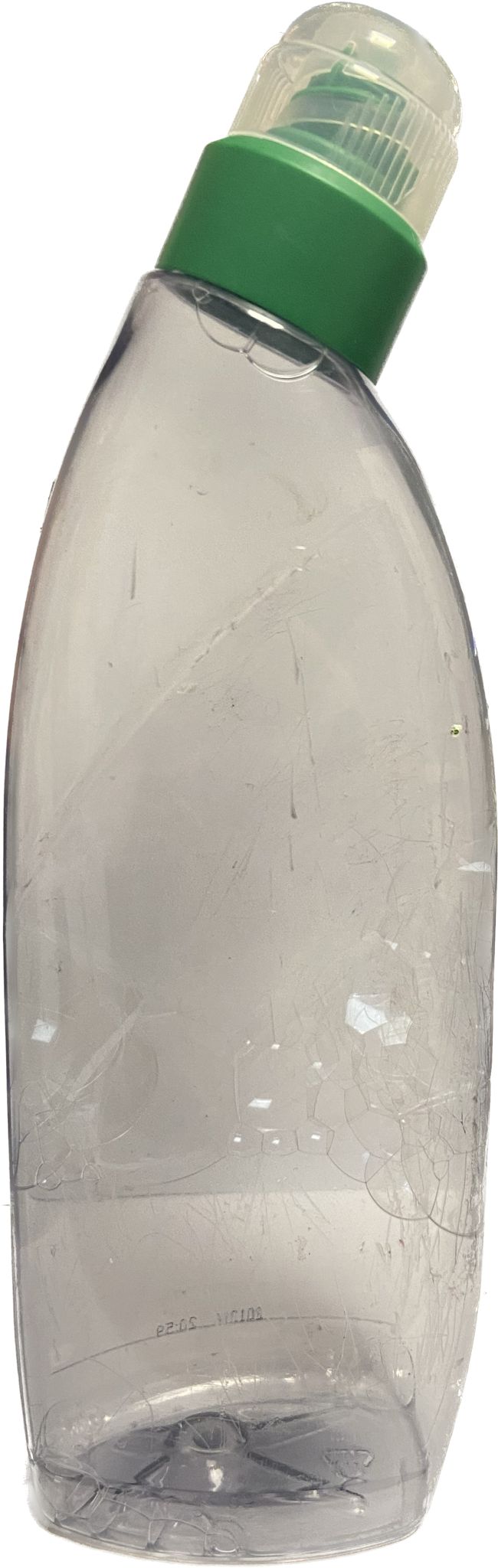}} &      \parbox[c]{1em}{
\includegraphics[height=1cm]{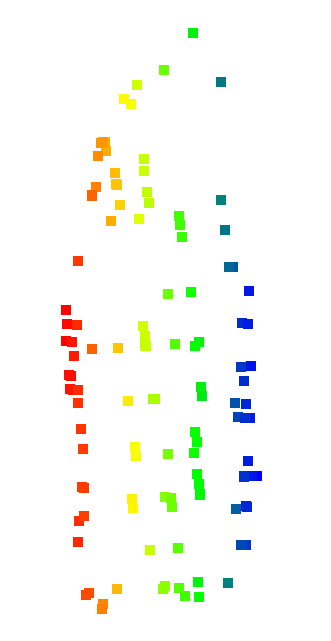}}    &   \parbox[c]{1em}{\includegraphics[height=1cm]{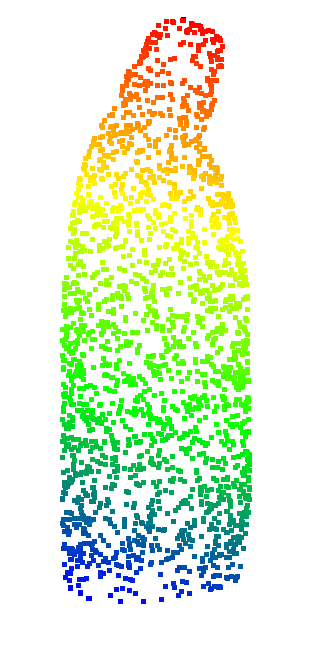}}         &      \parbox[c]{1em}{\includegraphics[height=1cm]{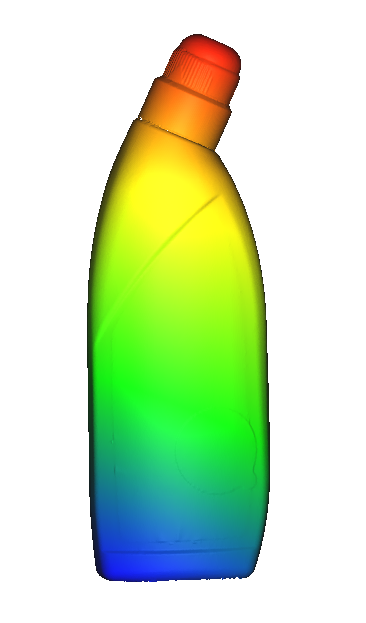}}                  &   \parbox[c]{1em}{\includegraphics[height=1cm]{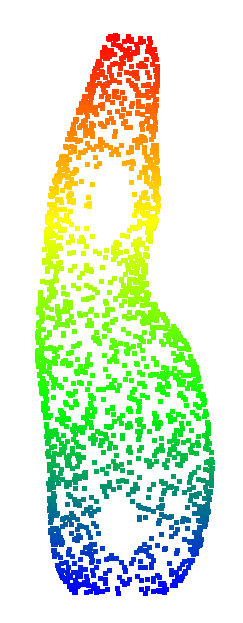}}        &    \parbox[c]{1em}{\includegraphics[height=1cm]{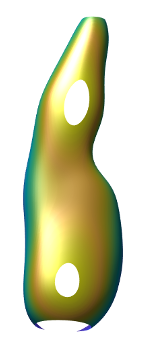}}            &   \parbox[c]{1em}{\includegraphics[height=1cm]{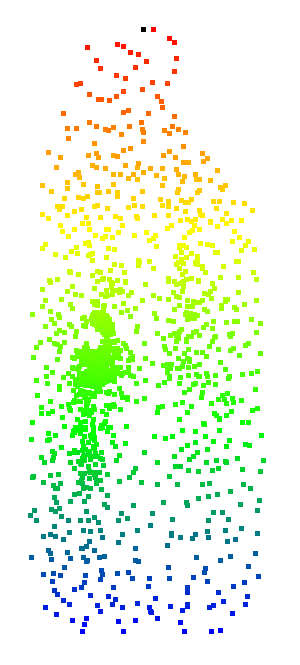}}           &    \parbox[c]{1em}{\includegraphics[height=1cm]{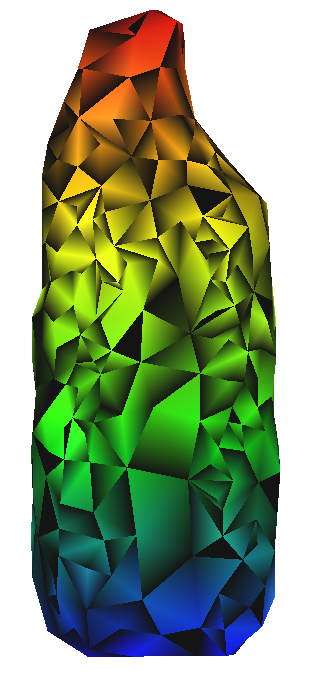}}                \\
Cup 1 &  \parbox[c]{1em}{
\includegraphics[height=0.8cm]{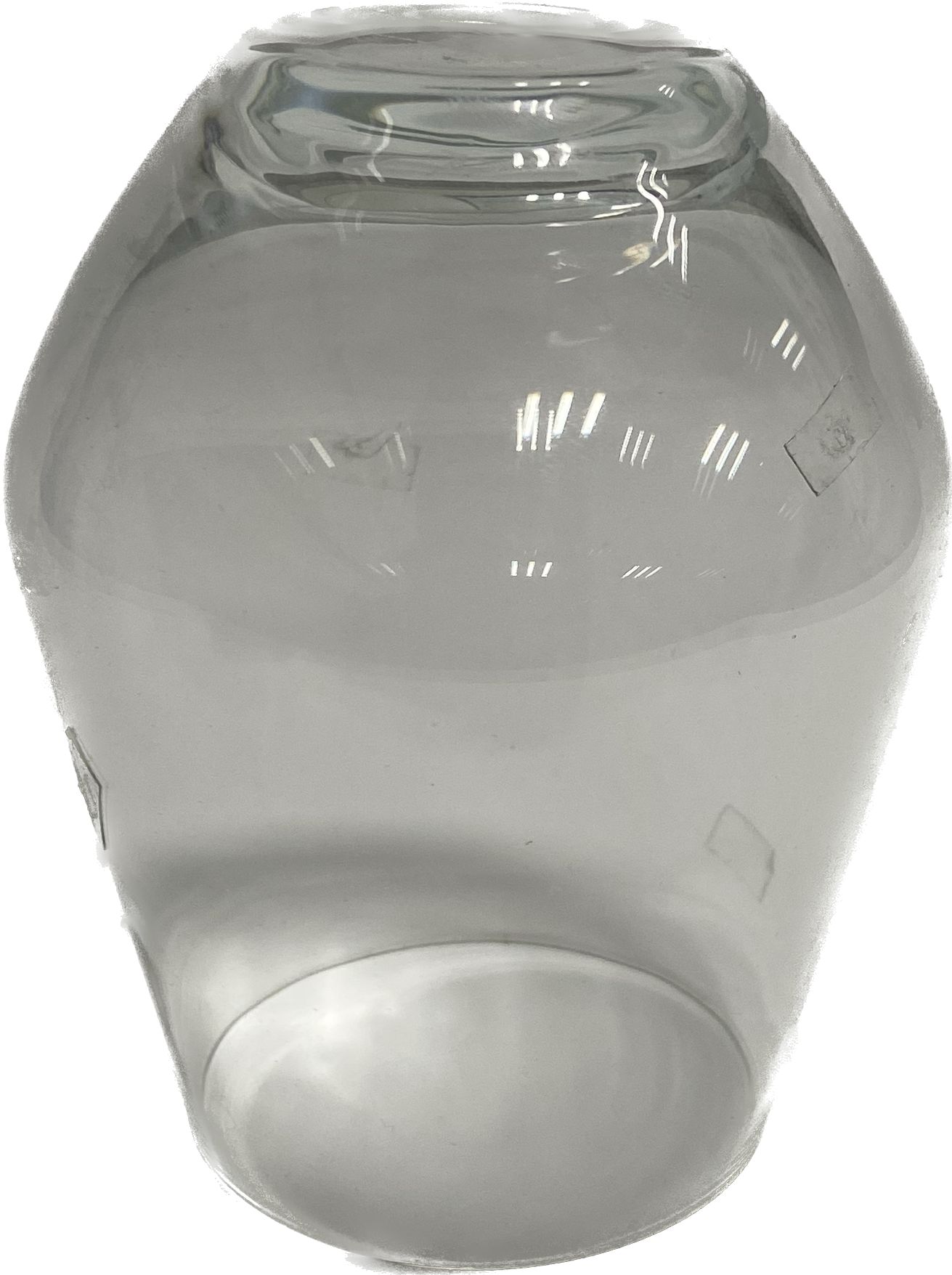}} \ \ &    \parbox[c]{1em}{
\includegraphics[height=1cm]{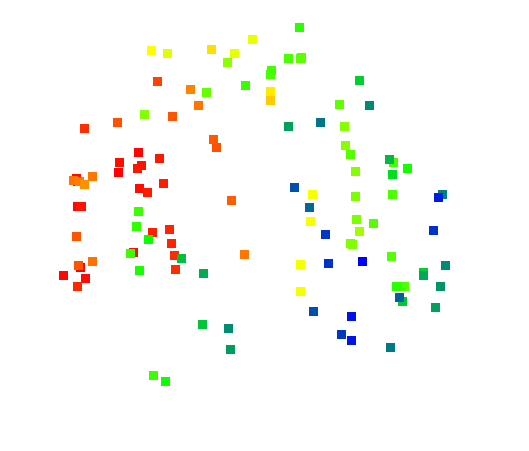}}      &      \parbox[c]{1em}{\includegraphics[height=0.9cm]{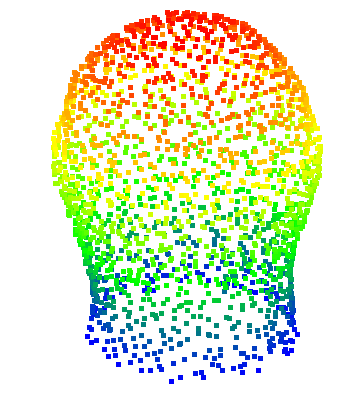}}         &  \parbox[c]{1em}{\includegraphics[height=1cm]{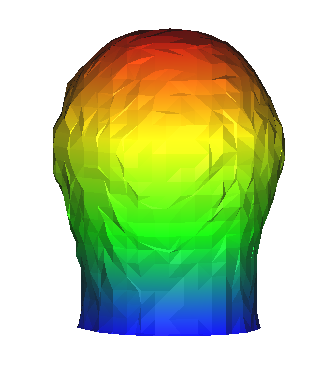}}                          &   \parbox[c]{1em}{\includegraphics[height=1cm]{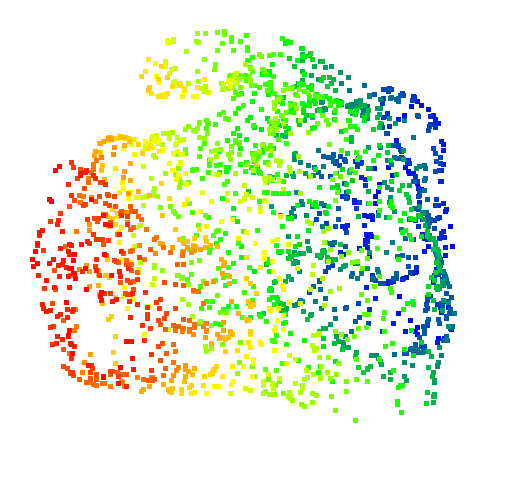}}        &  \parbox[c]{1em}{\includegraphics[height=1cm]{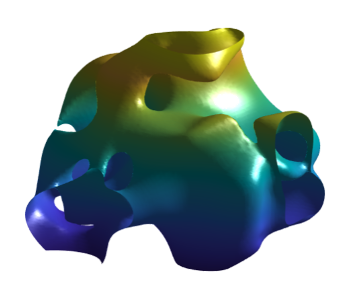}}              &     \parbox[c]{1em}{\includegraphics[height=1cm]{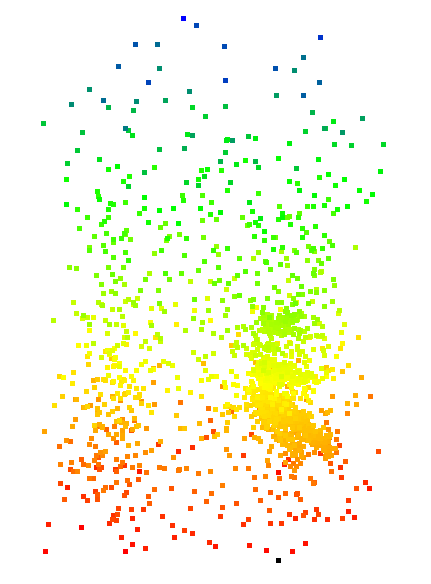}}           &        \parbox[c]{1em}{\includegraphics[height=1cm]{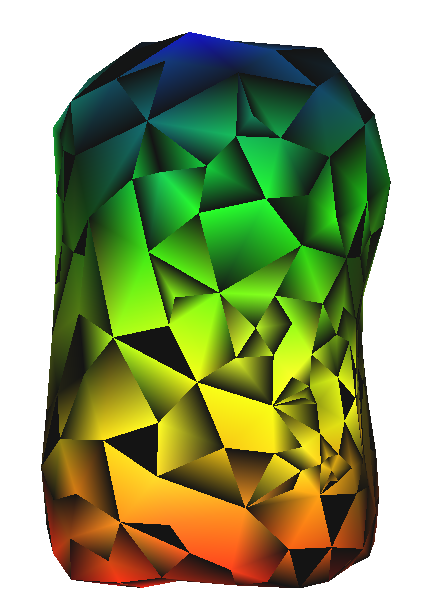}}            \\
Cup 2  &  \parbox[c]{1em}{
\includegraphics[height=1cm]{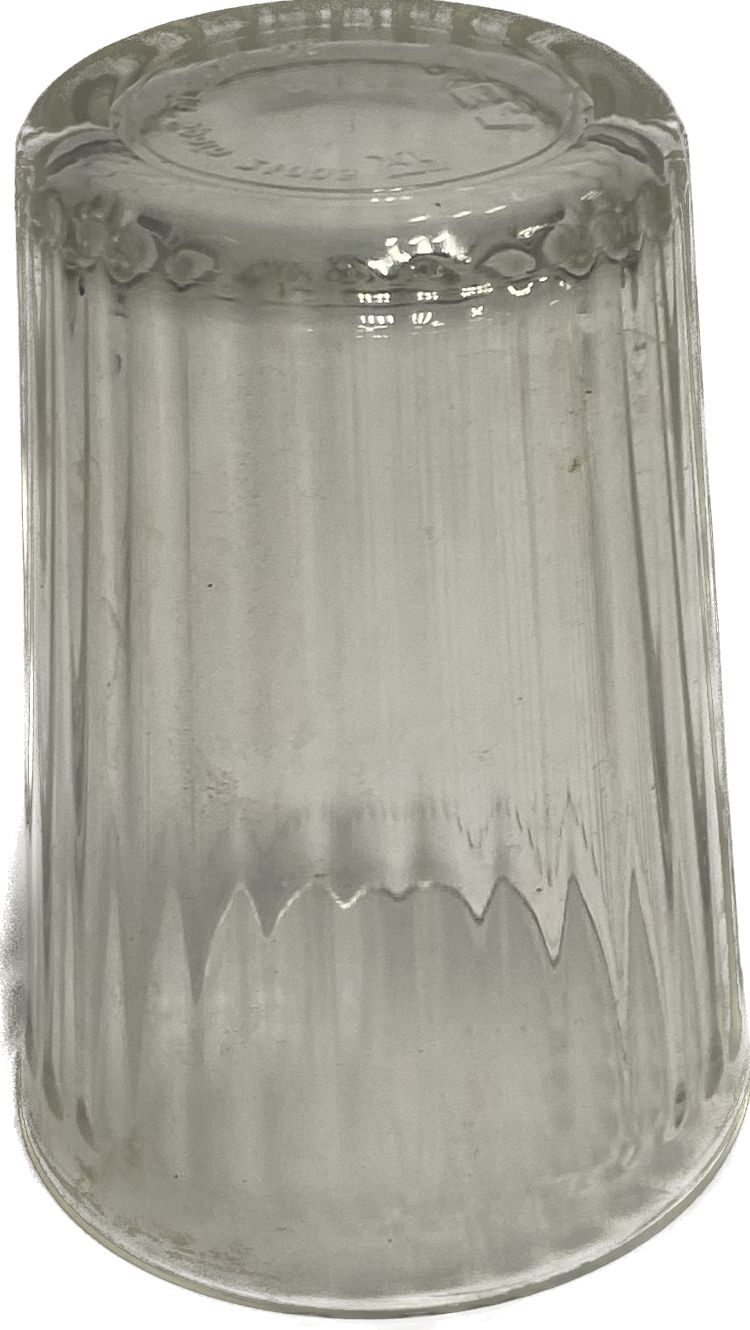}} &   \parbox[c]{1em}{
\includegraphics[height=1cm]{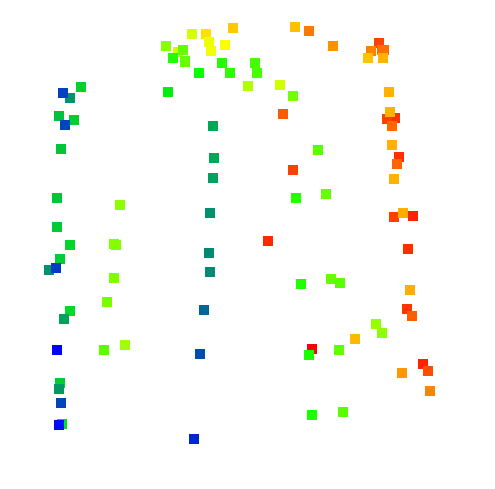}}     &     \parbox[c]{1em}{\includegraphics[height=0.9cm]{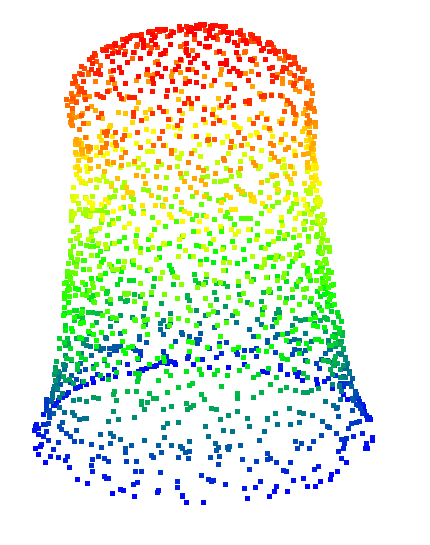}}          &    \parbox[c]{1em}{\includegraphics[height=1cm]{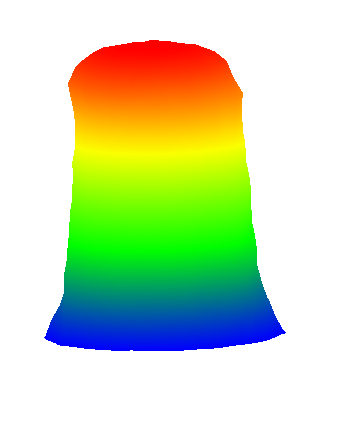}}                          &   \parbox[c]{1em}{\includegraphics[height=1cm]{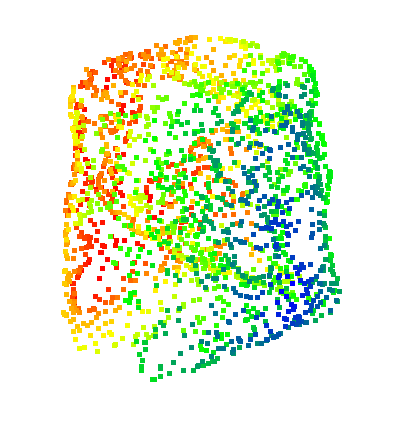}}        &      \parbox[c]{1em}{\includegraphics[height=1cm]{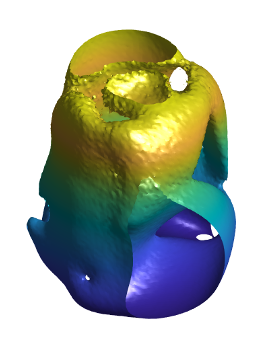}}          &   \parbox[c]{1em}{\includegraphics[height=1cm]{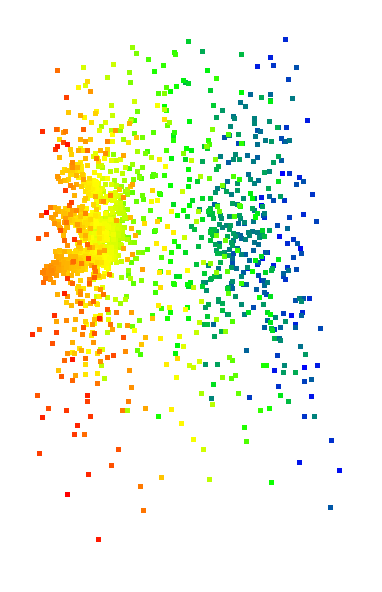}}           &        \parbox[c]{1em}{\includegraphics[height=1cm]{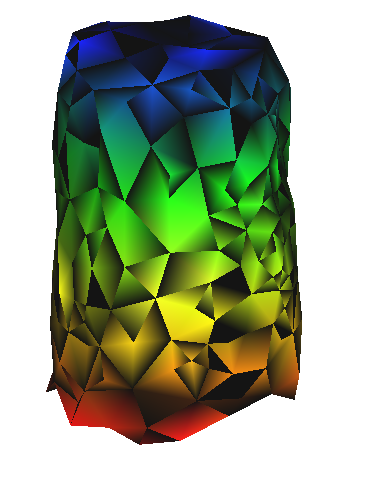}}          \\
Cup 3  &  \parbox[c]{1em}{
\includegraphics[height=1cm]{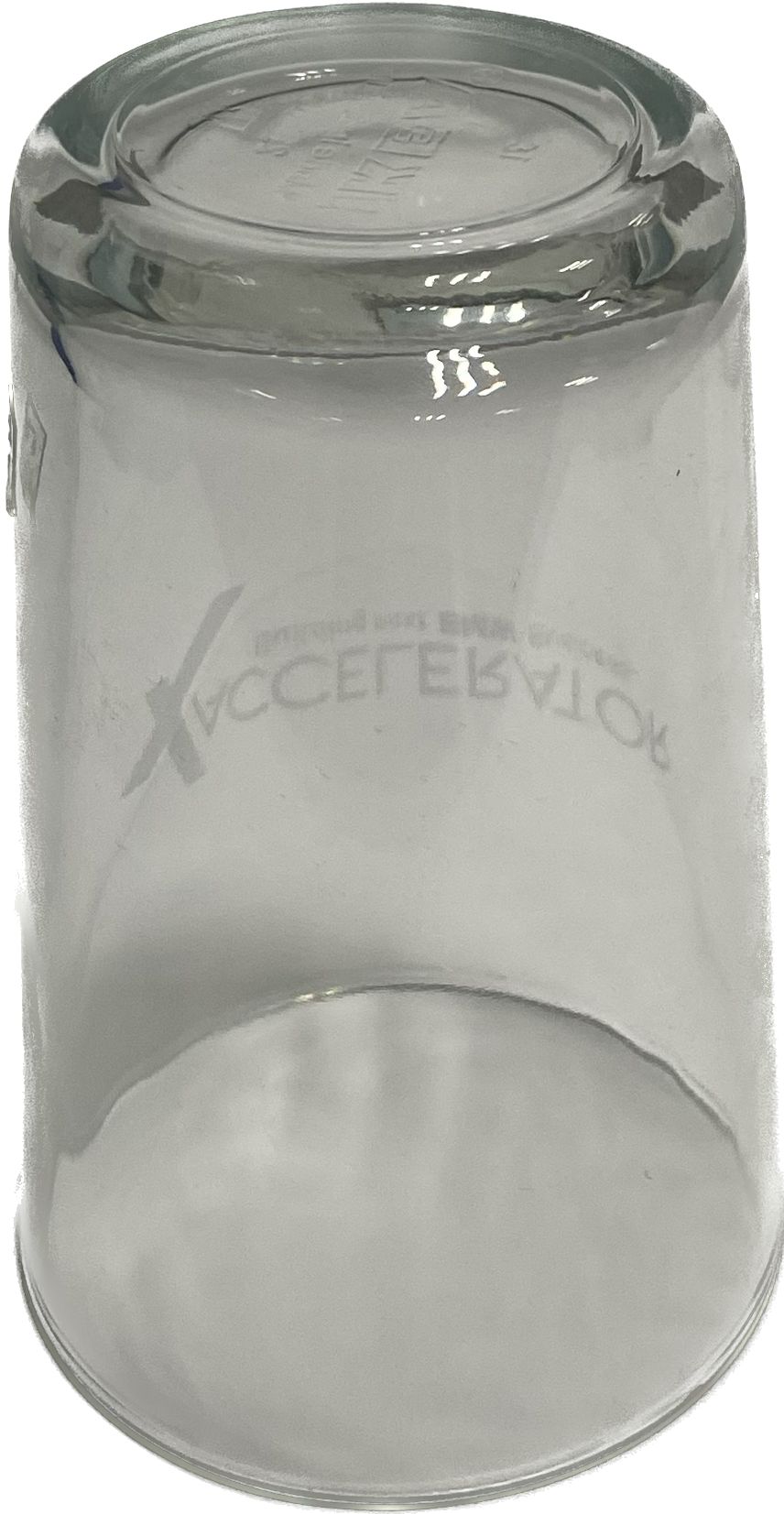}} &    \parbox[c]{1em}{
\includegraphics[height=1cm]{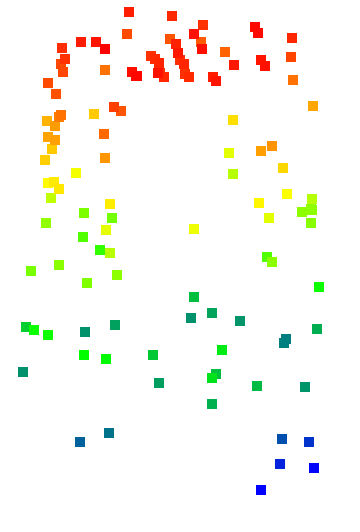}}             &     \parbox[c]{1em}{\includegraphics[height=1cm]{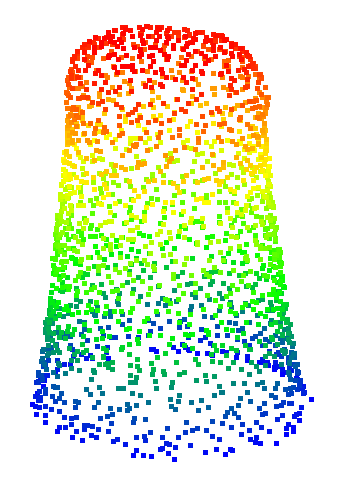}}          &  \parbox[c]{1em}{\includegraphics[height=1cm]{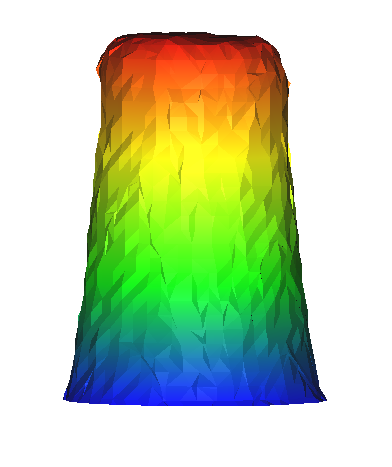}}                    &    \parbox[c]{1em}{\includegraphics[height=1cm]{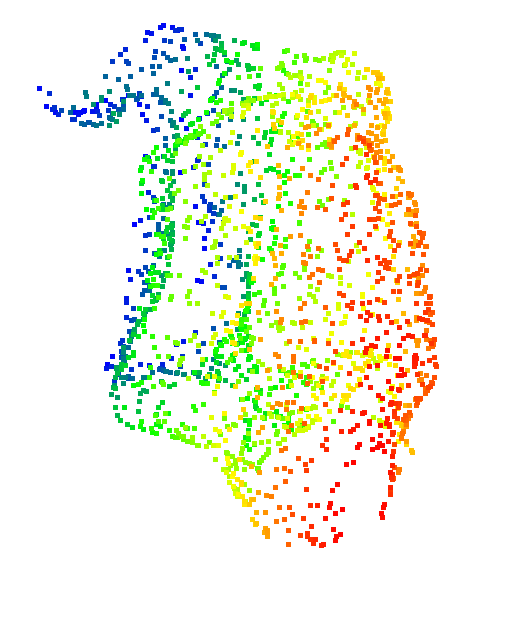}}       &     \parbox[c]{1em}{\includegraphics[height=1cm]{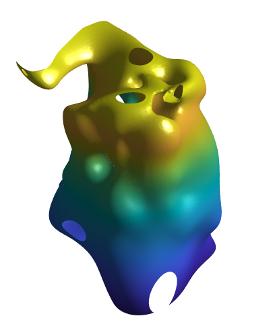}}           &     \parbox[c]{1em}{\includegraphics[height=1cm]{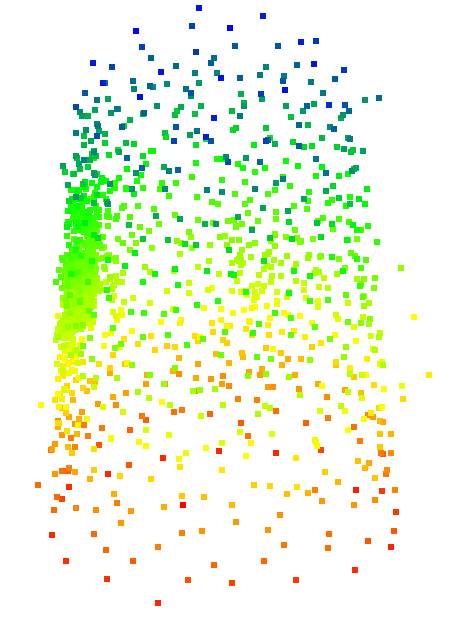}}         & \parbox[c]{1em}{\includegraphics[height=1cm]{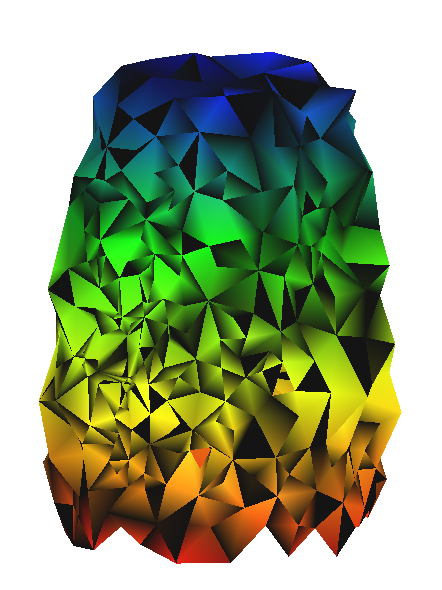}}                \\
Wineglass &\parbox[c]{1em}{
\includegraphics[height=1cm]{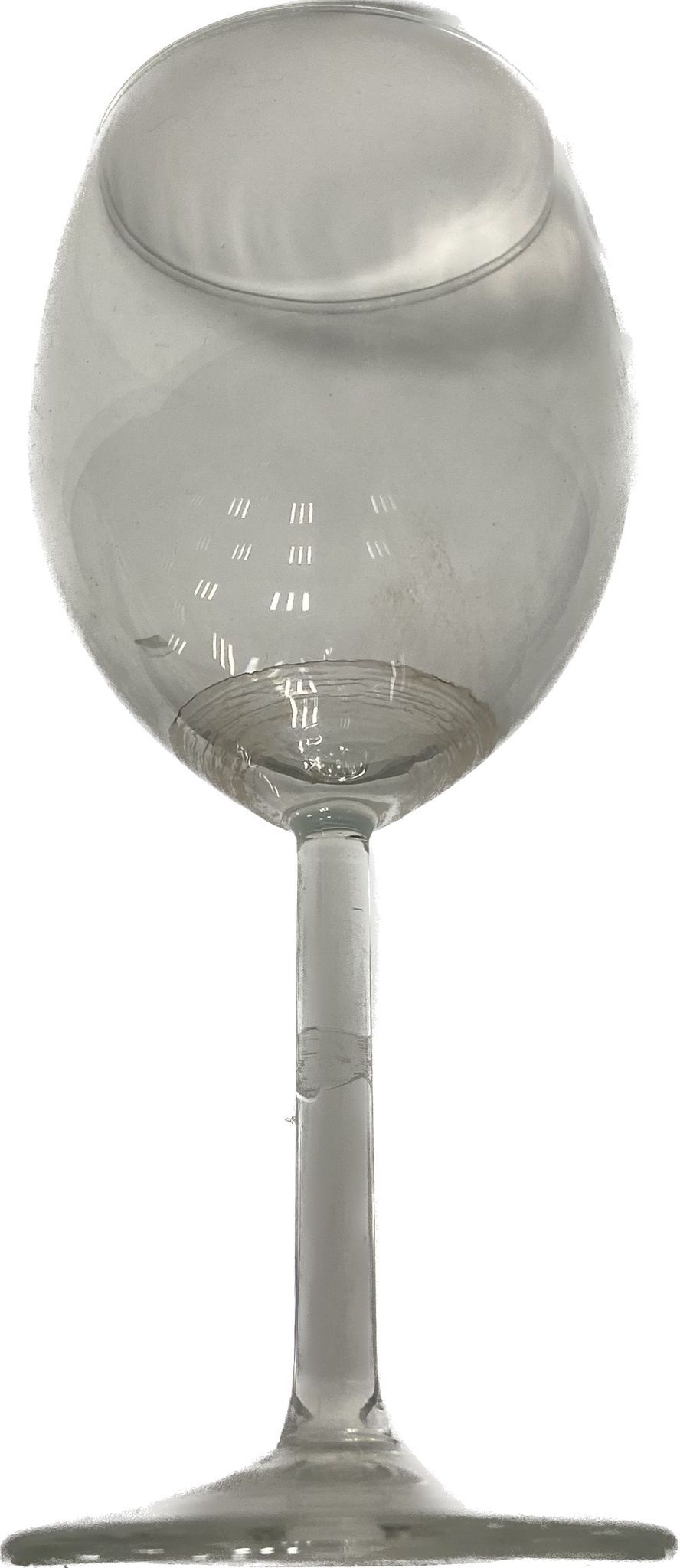}}&   \parbox[c]{1em}{
\includegraphics[height=1cm]{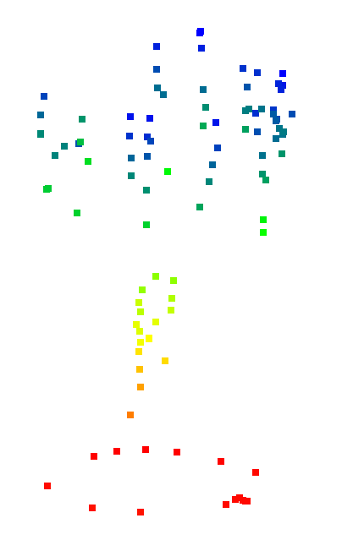}}        &     \parbox[c]{1em}{\includegraphics[height=0.9cm]{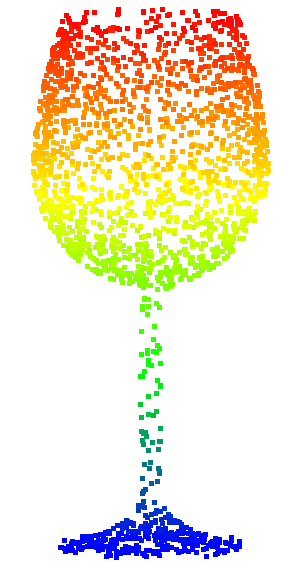}}          &   \parbox[c]{1em}{\includegraphics[height=1cm]{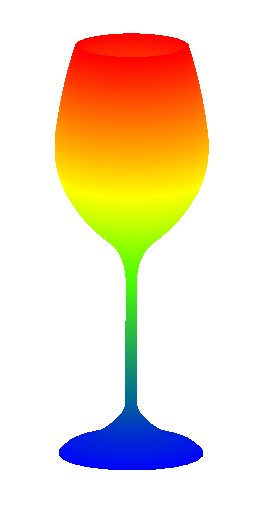}}                    &     \parbox[c]{1em}{\includegraphics[height=1cm]{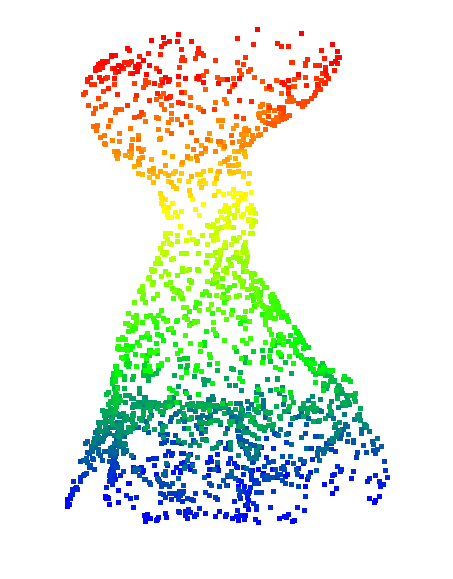}}      &     \parbox[c]{1em}{\includegraphics[height=1cm]{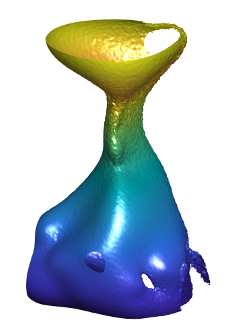}}           &     \parbox[c]{1em}{\includegraphics[height=1cm]{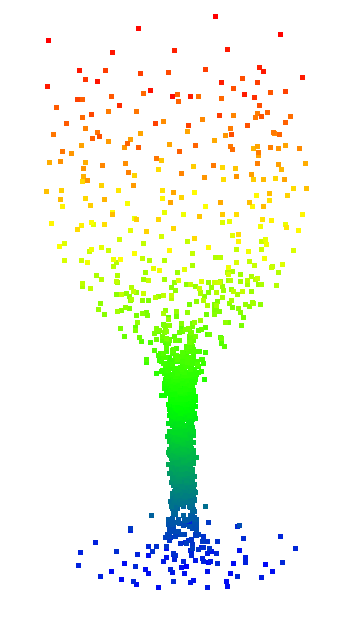}}         &            \parbox[c]{1em}{\includegraphics[height=1cm]{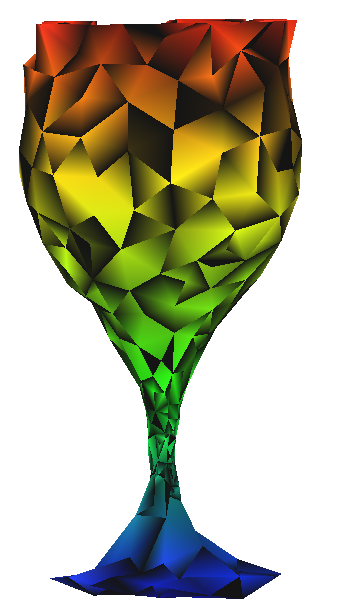}}          \\
Spray &  \parbox[c]{1em}{
\includegraphics[height=1cm]{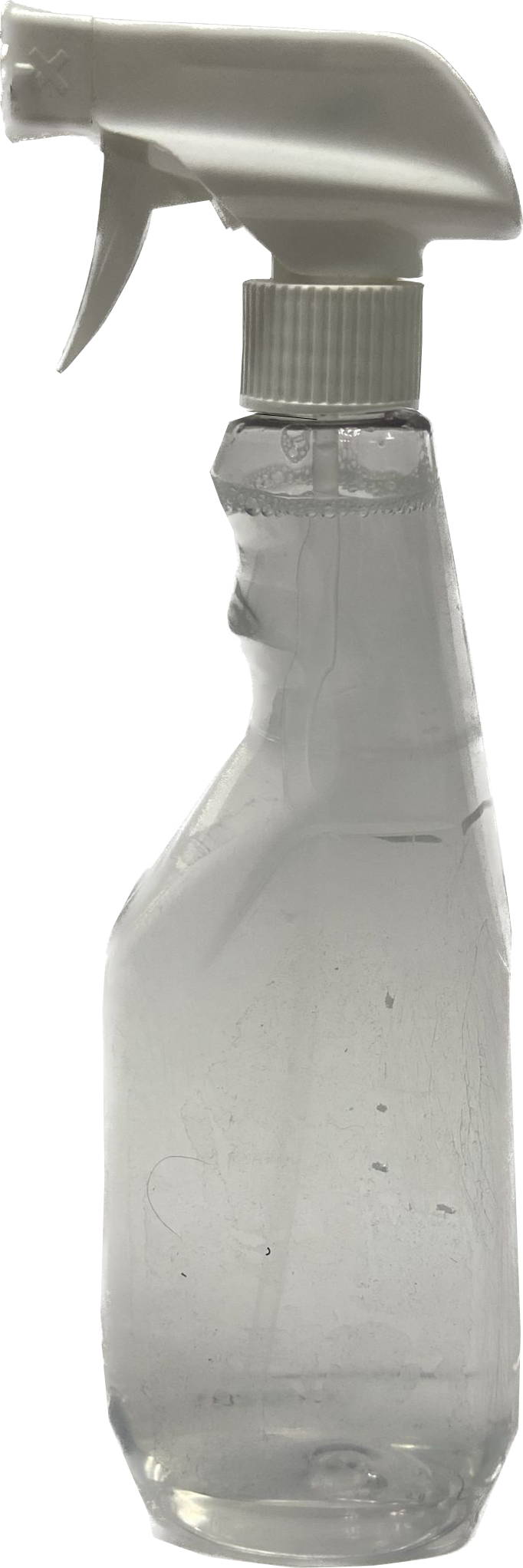}}  &      \parbox[c]{1em}{
\includegraphics[height=1cm]{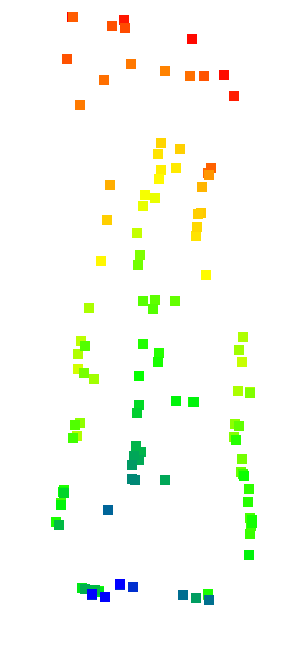}}     &    \parbox[c]{1em}{\includegraphics[height=1cm]{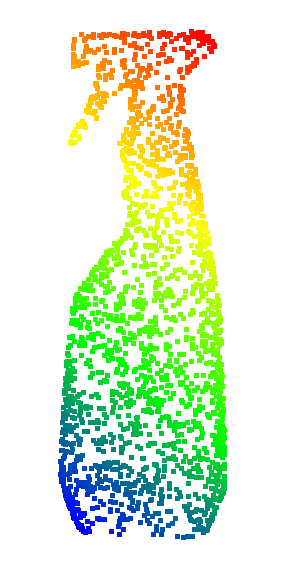}}           &    \parbox[c]{1em}{\includegraphics[height=1cm]{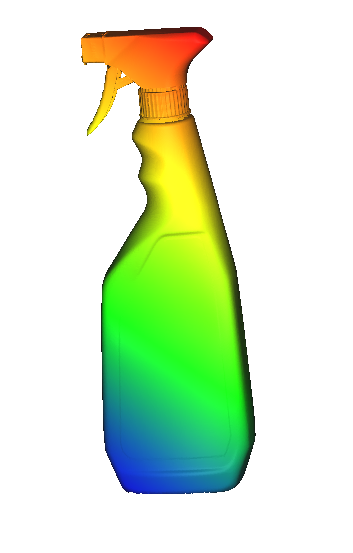}}                      &  \parbox[c]{1em}{\includegraphics[height=1cm]{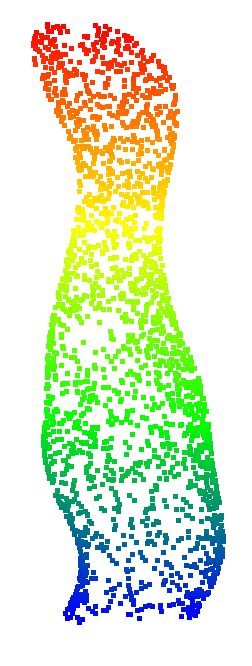}}         &     \parbox[c]{1em}{\includegraphics[height=1cm]{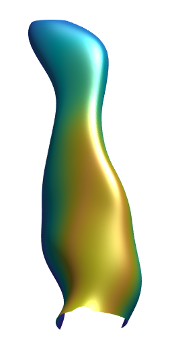}}           &    \parbox[c]{1em}{\includegraphics[height=1cm]{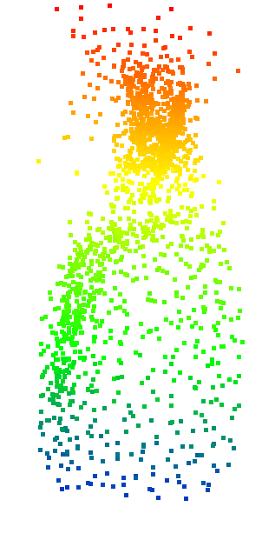}}           &  \parbox[c]{1em}{\includegraphics[height=1cm]{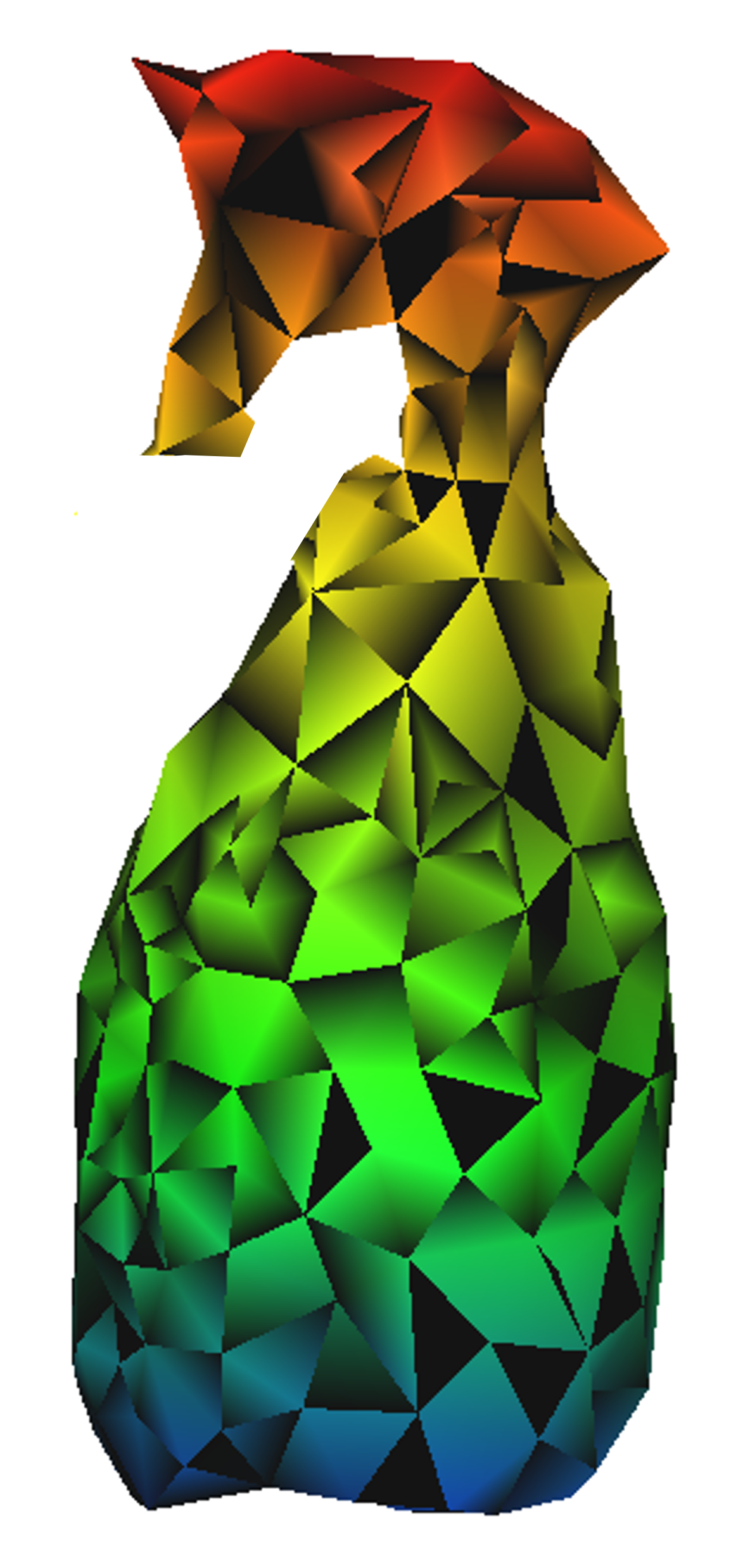}}                  \\ \bottomrule
\end{tabular}%
}
\end{table*}
\begin{figure*}[t!]
    \centering
    \begin{subfigure}[b]{0.32\textwidth}
    \includegraphics[width=\textwidth]{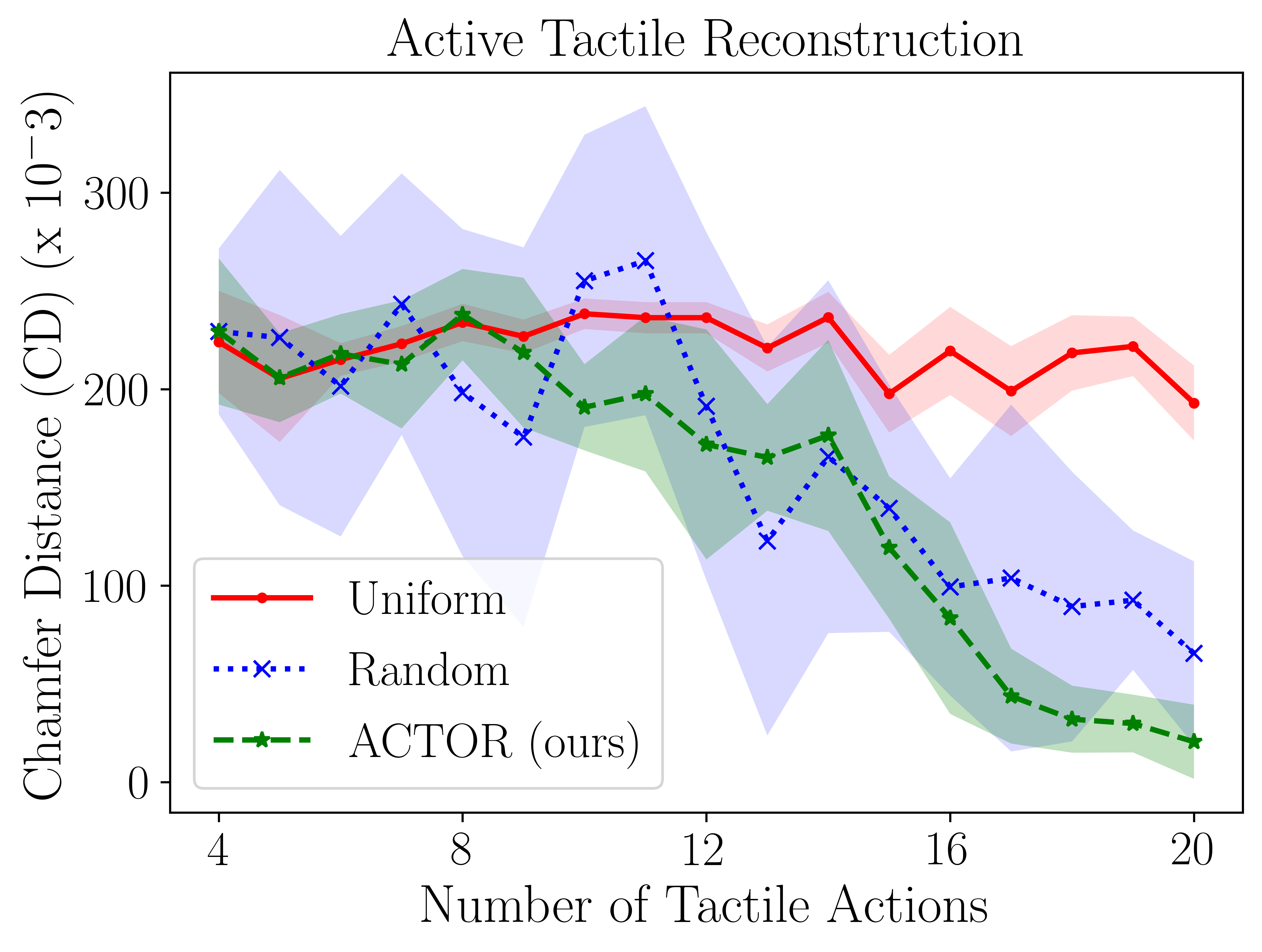}
    \caption{}
    \label{fig:active_tactile}
    \end{subfigure}
    \begin{subfigure}[b]{0.32\textwidth}
    \includegraphics[width=\textwidth]{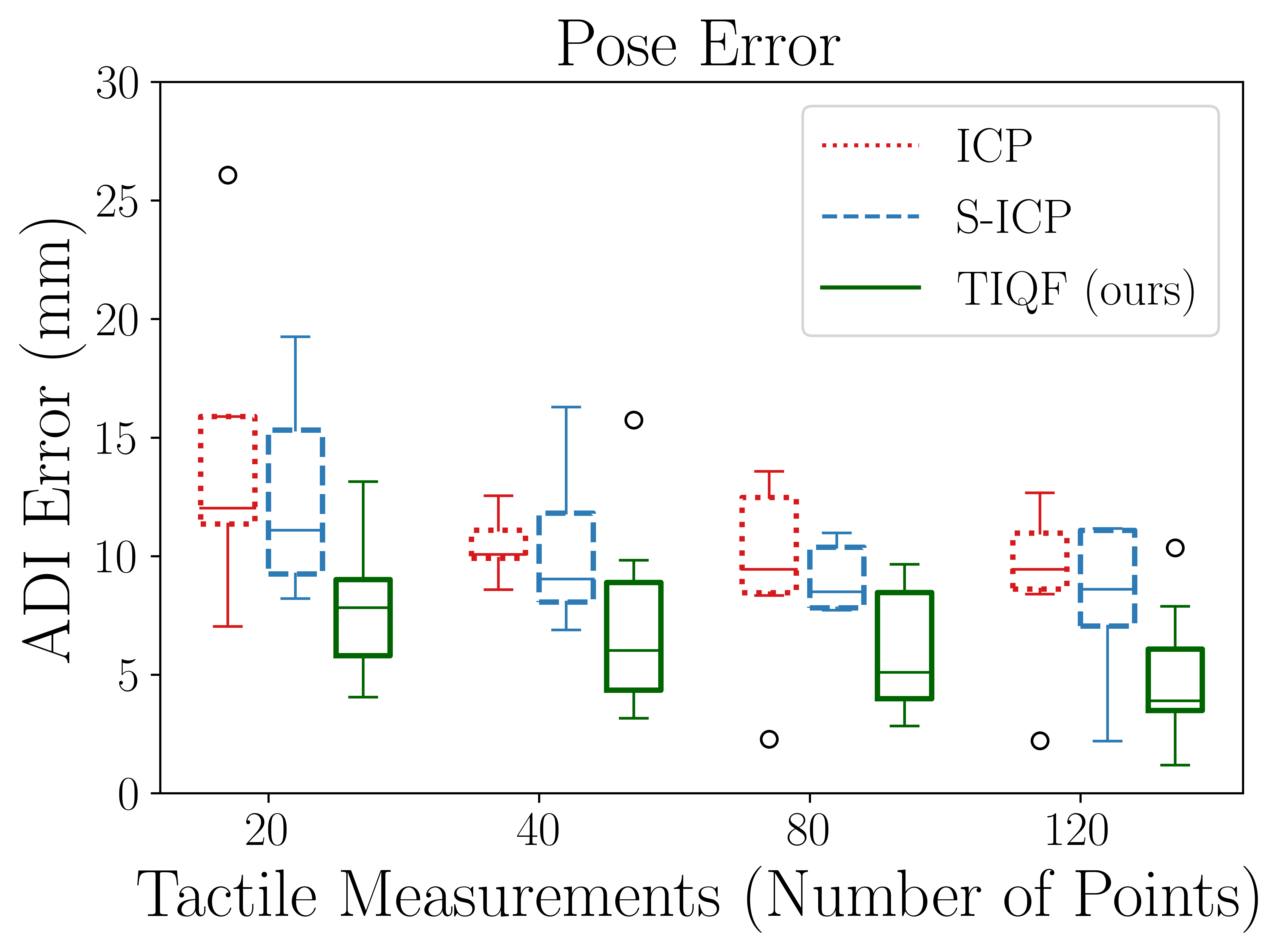}
    \caption{}
    \label{fig:pose_est}
    \end{subfigure}
    \begin{subfigure}[b]{0.32\textwidth}
    \includegraphics[width=\textwidth]{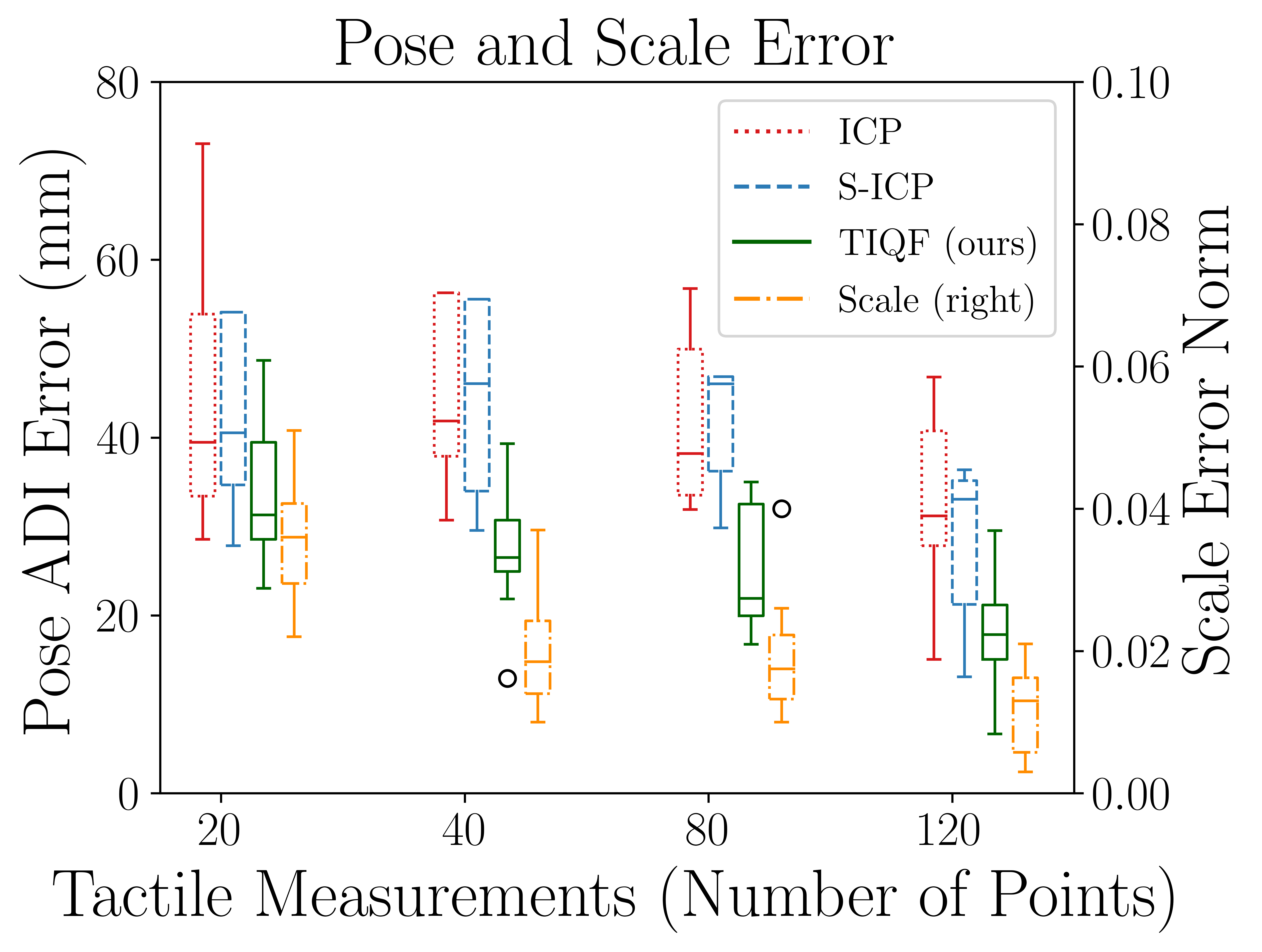}
    \caption{}
    \label{fig:pose_est_cat}
    \end{subfigure}
    \caption{(a) Active tactile reconstruction accuracy evaluated using the chamfer distance with ground-truth, (b) Pose estimation with ground truth object point cloud from CAD mesh, (c) Pose and scale estimation with reconstructed point cloud as object point cloud}
    \label{fig:active_and_plots}
\end{figure*}
The height of the occupancy grid is set constant for every object at 0.4m which is larger than the biggest object. 
Reconstruction with acceptable accuracy is obtained with 100 points or more as input. For each object, ten tactile point clouds with point number between 100 and 120 points are extracted using the active exploration strategy and used for reconstruction. The ground-truth point cloud and CAD mesh are obtained by spray-painting the objects and using a scanning device. For evaluation, we use the following performance metrics: Hausdorff distance (HD), Chamfer distance (CD) and Earth Mover distance (EMD). The CD is described in Sec.~\ref{ssec:deep_reconstruction}. Given two points $S_1$ and $S_2$, the Hausdorff distance is defined as~\cite{berger2013benchmark}:
\begin{equation}
    HD(S_1, S_2) = \max\{ \max_{x \in S_1}\min_{y \in S_2}\ \{ ||x - y ||_{2} \}, \max_{y \in S_2}\min_{x \in S_1} \{||y - x ||_{2} \} \}
    \label{eq:hd}
\end{equation}
The HD represents the maximum distance between the two point sets and can be affected by extreme outliers during the reconstruction.
The EMD finds a bijection $\phi: S_1 \rightarrow S_2$ to minimise the average distance between corresponding points in the point clouds as:
\begin{equation}
    EMD(S_1, S_2) = \min_{\phi: S_1 \rightarrow S_2} \frac{1}{|S_1|}\sum_{x \in S_1}  || x - \phi(x)||_2 \quad .
    \label{eq:emd}
\end{equation}
A perfect reconstruction will yield $\{CD, HD, EMD\} \rightarrow 0$ and lower values signify better reconstruction.  \\
We use Gaussian Process Implicit Surfaces (GPIS) as baseline as it is widely used in the literature for tactile-based object reconstruction~\cite{dragiev2011gaussian, yi2016active, bjorkman2013enhancing, gandler2020object, martens2016geometric, suresh2021tactile, jamali2016active}. For implementation, we utilise the GP for machine learning toolbox~\cite{rasmussen2010gaussian} in MATLAB and the Mat\'ern kernel.

The quantitative results of tactile-based reconstruction using our method and baseline GPIS method are shown in Fig.~\ref{fig:quant_plots} and qualitative reconstruction results are presented in Tab.~\ref{tab:qualitative_results}.
From Fig.~\ref{fig:quant_plots}, we note our proposed approach yields lower CD values for all objects. For HD and EMD, apart from the bottle and spray, our method performs better than the baseline approach. On average, our approach is 45\%, 23.5\% and 28\% lower in CD, HD and EMD values compared to baseline GPIS. While the quantitative results focus on local point-distances between the reconstructed and ground-truth point cloud, the qualitative results in Tab.~\ref{tab:qualitative_results} demonstrate the differences in reconstruction accuracy at the object level. GPIS produces warped reconstructed surfaces due to the low number of tactile points. Whereas our method, with the help of the learned model over the category-level synthetic objects, is able to reconstruct the object to an acceptable accuracy even with sparse input data.

\textbf{Active Tactile Reconstruction:} Using our proposed framework ACTOR, we can achieve accurate reconstruction with fewer tactile actions in comparison to the baselines as shown in Fig.~\ref{fig:active_tactile}. 
We define an uniform object exploration and random object exploration strategy as baselines as follows: the bounding box around the object is transformed into a grid with each grid cell of size 3cm $\times$ 3cm (size of the sensor patch). The grid does not encode the probabilistic occupancy as in our ACTOR approach. The robot explores each grid cell in a sequential manner in the uniform strategy. In contrast, for the random strategy, the robot picks a grid cell at random for exploration.
In order to have an unbiased comparison between the exploration methods, a maximum of 20 actions are chosen as on average it takes 20 actions to extract atleast 100 tactile points. We begin the model inference from the 4th action onwards to have a minimum of 20 points in the tactile point cloud.
We note that the uniform strategy requires a large number of tactile actions to completely explore the object in order for reconstruction. The random  strategy has high variance in terms of reconstruction accuracy and stems from the stochastic nature of the exploration while ACTOR deterministically improves reconstruction accuracy with the increasing number of tactile actions. 



\subsection{Tactile-based Transparent Object Pose Estimation}
As the error in reconstruction propagates to downstream tasks, we perform two experiments: firstly, instance-level estimation using the ground-truth model point cloud as the object point cloud (Fig.~\ref{fig:pose_est}) and secondly, category-level pose estimation using the reconstructed point cloud as the object point cloud (Fig.~\ref{fig:pose_est_cat}). For category-level pose estimation, norm scale error is also reported in addition to rotation and translation.
As our proposed TIQF method is a local registration method, we chose the standard Iterative Closest Point (ICP)~\cite{besl1992method} and Sparse Iterative Closest Point (S-ICP)~\cite{bouaziz2013sparse} as baselines. S-ICP is chosen as it demonstrates higher robustness to outliers and incomplete data as typically found in tactile point clouds. We use the Average Distance of model points with Indistinguishable views metric (ADI)~\cite{hinterstoisser2013model} as a combined measure of the rotation and translational error as we have multiple objects with axis of symmetry. The ADI metric is defined as:    
\begin{equation}
    \mathtt{err}_{adi} = \frac{1}{|\mathcal{O}|}\sum_{\mathbf{p}_1 \in \mathcal{O}} \min_{\mathbf{p}_2 \in \mathcal{O}} || (\mathbf{R}_{gt}\mathbf{p}_1 + \mathbf{t}_{gt}) - (\mathbf{R}_{est}\mathbf{p}_2 + \mathbf{t}_{est}) ||,
    \label{eq:adi}
\end{equation}
where $(\mathbf{R}_{gt}, \mathbf{t}_{gt})$ and $(\mathbf{R}_{est}, \mathbf{t}_{est})$ refers to ground-truth and estimated rotation and translation respectively.
As seen from Fig.~\ref{fig:pose_est},\ref{fig:pose_est_cat}, our proposed approach outperforms the baseline approaches for all input tactile point clouds with varying point numbers demonstrating robustness to point sparsity. The median $\mathtt{err}_{adi} < 1cm$ for our proposed approach even with sparse point clouds with  $N_{P^t} = 20$  and improves with increasing the number of points. The category-level pose estimation errors are higher than instance-level due to the errors in the reconstructed point clouds. However, the accuracy improves by reducing scale error with median $ \mathtt{err}_{adi} < 2cm$ for $N_{P^t} = 120$ with our proposed method.

\subsection{Discussion}
Our proposed approach, ACTOR outperforms the GPIS strategy by all our evaluation metrics. We also note the qualitative reconstruction results in Tab.~\ref{tab:qualitative_results}, wherein GPIS fails to capture the shape details of the object while our approach captures the global and local shape accurately (see object spray and wineglass). Our network implicitly learns important feature points and is able to reconstruct the object accurately given few sparse inputs.
Our active exploration strategy converges faster to reconstruct the object shape thus improving the sample efficiency.
Furthermore, our proposed category-level pose estimation method outperforms the baseline methods by $\geq$ 25\% ADI error. 

A limitation of the work is the need for category-wise object models for training. A possible future work includes using neural radiance fields (NeRFs)~\cite{wang2021nerf} to generate synthetic models of objects from images that can be used for training.

\section{CONCLUSIONS}
\label{sec:conclusions}
In this work we proposed ACTOR, a novel framework for active tactile-based category-level transparent object reconstruction. By learning with only synthetic object models, ACTOR is capable of performing real-world transparent object reconstruction through sparse tactile data. Our approach outperforms state-of-the-art GPIS method in terms of reconstruction accuracy. Furthermore, we demonstrated category-level pose estimation with the reconstructed object model and our approach outperforms baseline ICP and S-ICP methods.
As future work, we would like to extend ACTOR for safely manipulating transparent objects in unstructured scenarios with possible deformability and dynamic center of mass~\cite{kaboli2016tactile, kaboli2015hand, yao2017tactile}.


\bibliography{root}
\bibliographystyle{IEEEtran}

\end{document}